\documentclass[]{fairmeta}

\usepackage{amsmath,amssymb,amsthm}
\usepackage{graphicx}
\usepackage{booktabs}
\usepackage{colortbl}
\usepackage{multirow}
\usepackage{array}
\usepackage{colortbl}
\usepackage{tabularx}
\usepackage{subcaption}
\usepackage{enumitem}
\usepackage{pifont}   
\usepackage{xspace}
\usepackage{lmodern}
\usepackage{eso-pic}
\usepackage{wrapfig}
\usepackage{placeins}

\usepackage[numbers,sort&compress]{natbib}
\usepackage[colorlinks=true,breaklinks=true,
            linkcolor=metablue,citecolor=metablue,urlcolor=metablue]{hyperref}
\ifdefined\pdfsuppresswarningpagegroup\pdfsuppresswarningpagegroup=1\fi

\setlist{leftmargin=*, itemsep=0.5em}

\newtcolorbox{keybox}[1][]{colback=metabg,colframe=metablue,
  boxrule=0.6pt,arc=6pt,left=6pt,right=6pt,top=6pt,bottom=6pt,#1}

\definecolor{metablue}{HTML}{26527A}

\logo{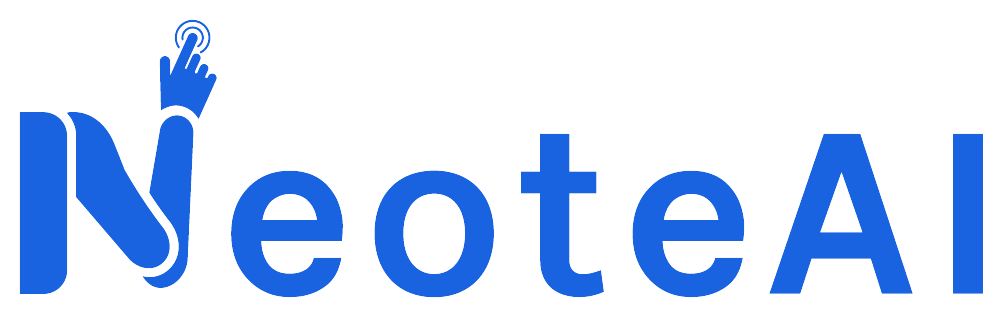}   
\logowidth{3cm}
\toplogo{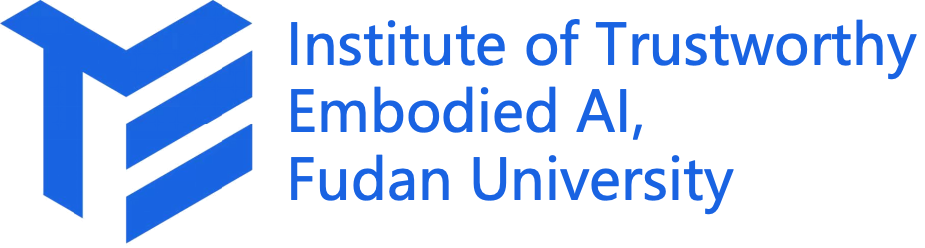}   
\toplogowidth{3.0cm}

\title{\texorpdfstring{$\mathcal{N}_0$}{N0}-Foundation: \\Towards the Age of Tactile Intelligence}

\author{NeoteAI Team \& Fudan TEAI Team}

\date{\today}
\metadata[Website]{\url{http://research.neoteai.com/n0-foundation}}
\metadata[Dataset]{\url{https://huggingface.co/datasets/NeoteAIEmbodied/OpenNeoData}}

\begin{document}

\abstract{We present $\mathcal{N}_0$-Foundation, a paradigm for tactile-enabled embodied manipulation, which integrates tactile sensing hardware, large-scale multimodal data, tactile representation learning, and standardized evaluation.
First, we engineer the underlying tactile infrastructure for scalable data collection and practical deployment, including a vision-based tactile sensor, a Tactile Universal Manipulation Interface ($\mathcal{N}_0$-TacUMI), and a synchronized visual–tactile data collection system supporting both multiple robot embodiments and $\mathcal{N}_0$-TacUMI-based demonstrations.
Leveraging this infrastructure, we construct NeoData, which contains more than $30{,}000$ hours of synchronized visual and tactile demonstrations, spanning six embodiments, $450$ tasks, and billions of paired RGB and tactile frames collected through a mixture of real-robot teleoperation and $\mathcal{N}_0$-TacUMI demonstrations.
To facilitate open research, we further release OpenNeoData, a $5{,}000$-hour open-source subset of NeoData that spans six embodiments, more than $250$ tasks, and over $200$ skills. The dataset addresses a central limitation of existing manipulation corpora: contact states, local forces, and incipient slip are rarely observable from vision alone, yet they are critical for deformable-object manipulation, precise assembly, delicate force control, and sustained surface interaction.
Capitalizing on the large-scale, heterogeneous tactile measurements within NeoData, we propose NeoForce, a visuo-tactile representation learning model that uses dense three-axis force fields as unified, hardware-agnostic supervision to learn transferable tactile representations across different sensor designs for downstream embodied policies.
To enable systematic evaluation of tactile embodied models built upon our infrastructure, datasets and tactile representations, we further propose a comprehensive benchmark, which combines the real-world NeoReal suite and the simulated NeoSim suite for standardized evaluation of contact-rich manipulation. Experiments across both suites show that policies benefit from the physical contact state rather than from the device-specific appearance of the tactile signal. These results indicate that large-scale tactile data and supervision make a universal tactile representation possible, which in turn enables embodied policies to perform fine-grained contact-rich manipulation. We release the dataset, the representation, and the benchmark, aiming at supporting future work on tactile-enabled embodied manipulation.}
\maketitle


\section{Introduction}

\begin{figure}[t]
    \centering
    \includegraphics[width=\linewidth]{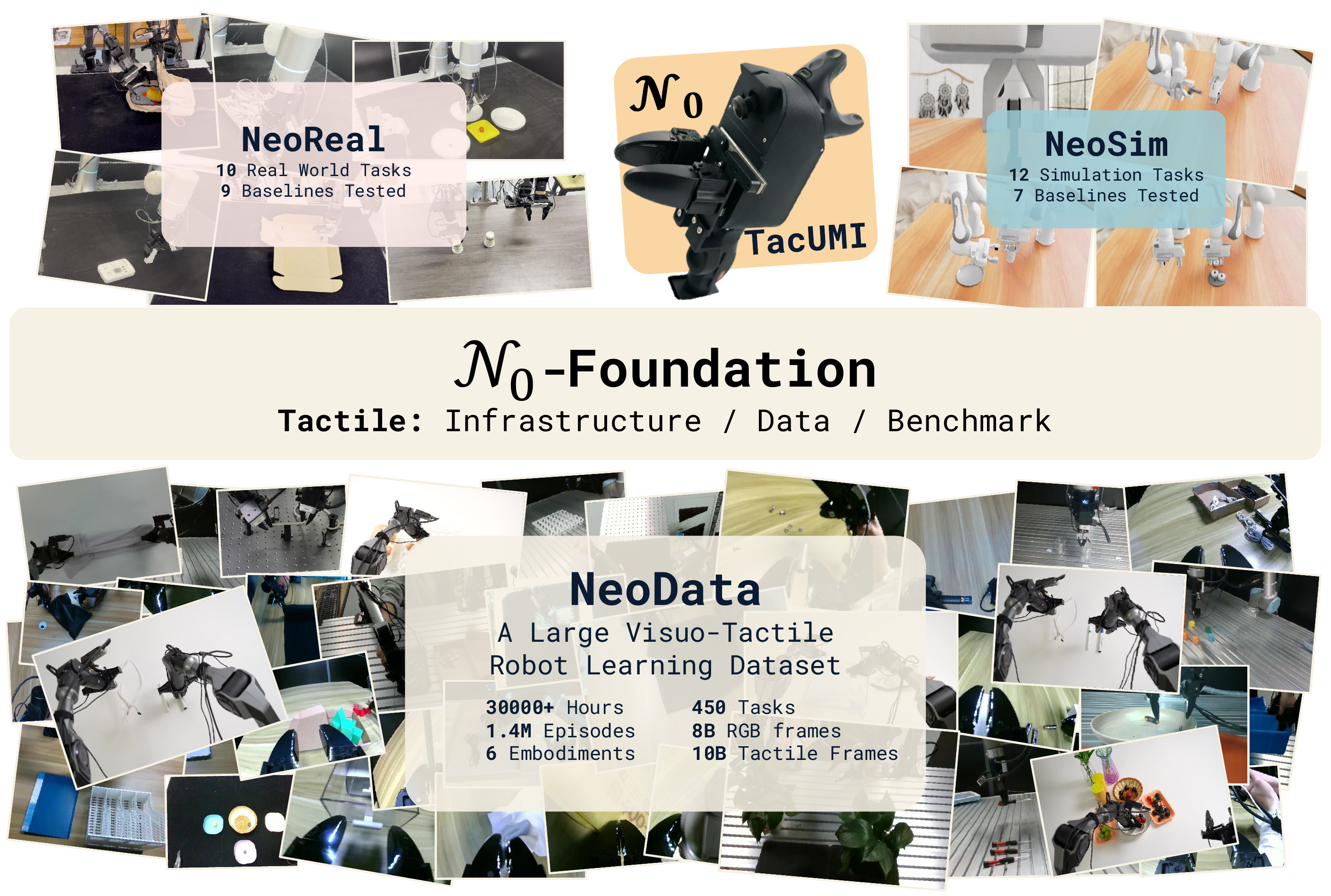}
    \caption{\textbf{$\mathcal{N}_0$-Foundation}, a tactile-centric foundation for embodied manipulation that unifies hardware, data, tactile representation and evaluation. At its core is NeoData, a large-scale visuo-tactile robot learning dataset with more than $30{,}000$ hours of interaction, $450$ tasks, $1.4$+M episodes, $8$B RGB frames, and $10$B tactile frames gathered from a mixture of real-robot embodiments and $\mathcal{N}_0$-TacUMI collection. On top of the data we evaluate tactile-based policies on the proposed real-world NeoReal suite and the simulated NeoSim suite.}
    \label{fig:teaser}
\end{figure}

Learning generalizable manipulation policies from large-scale demonstration data has become a central paradigm in embodied intelligence. Recent progress in vision-language-action (VLA) models has demonstrated that scaling diverse multimodal demonstrations substantially improves the generalization and robustness of visuomotor policies \citep{brohan2022rt1, oneill2024open, black2025pi05}. Complementing direct policy learning, World Action Models learn action-conditioned environment dynamics by modeling how the future world evolves under robot actions, showing promising generalization across real-world tasks and environments \citep{wu2024unleashing, li2025unified, ye2026dreamzero}. Progress in both paradigms is increasingly driven by the scale and diversity of available manipulation data.
Real-robot datasets range from teleoperated single-embodiment corpora \citep{walke2023bridgedata, mandlekar2021matters} to cross-embodiment collections gathered across multiple robots and institutions \citep{oneill2024open, khazatsky2024droid}. Portable collection systems such as the UMI \citep{chi2024universal} further enable scalable in-the-wild demonstration collection beyond fixed robot workcells and have inspired a growing family of embodiment-agnostic interfaces \citep{liu2024fastumi, xu2025dexumi, ha2024umi, zeng2025activeumi}. Meanwhile, large-scale egocentric videos provide abundant observations of human–object interactions, offering complementary behavioral and semantic priors for embodied learning \citep{damen2018epickitchens, hoque2026egodex, punamiya2026egoverse}.

Despite this progress, two limitations remain. First, most large-scale embodied datasets are dominated by visual observations. However, vision alone is often insufficient for contact-rich manipulation, where success depends on local force, friction, incipient slip, and contact transitions that are weakly observable from external or wrist-mounted cameras. These limitations affect a broad range of task families, including deformable-object manipulation, precise insertion and assembly, delicate force control, sustained surface interaction, and bimanual coordination. Second, the tactile modality remains fragmented. Camera-based visuo-tactile sensors \citep{yuan2017gelsight, lambeta2020digit}, capacitive arrays \citep{schmitz2011methods, maiolino2013flexible}, and fiber-based sensors \citep{xing2025taccap} produce signals in different hardware-specific formats. This heterogeneity prevents tactile demonstrations collected on one device from being reused directly on another, which complicates policy training over pooled tactile data and ultimately prevents tactile datasets and models from benefiting from scale.

To address these limitations, $\mathcal{N}_0$-Foundation, pronounced as Neo-Foundation, establishes a tactile-centric foundation for embodied manipulation. As summarized in Fig.~\ref{fig:teaser}, $\mathcal{N}_0$-Foundation establishes a scalable paradigm for tactile-enabled embodied learning that connects scalable tactile data infrastructure, large-scale multimodal data acquisition, unified representation learning, and standardized policy evaluation. $\mathcal{N}_0$-Foundation is organized around four technical pillars: tactile infrastructure, a large-scale dataset named NeoData, tactile representation learning with NeoForce, and a comprehensive evaluation protocol.
First, we build the underlying tactile infrastructure required for scalable collection and practical deployment. It includes durable vision-based tactile sensors, $\mathcal{N}_0$-TacUMI, and a synchronized visual--tactile data collection system supporting both real-robot demonstrations across multiple embodiments and portable UMI-based demonstrations. Built on this infrastructure, NeoData contains more than $30{,}000$ hours of synchronized visual and tactile demonstrations, spanning six embodiments, $450+$ tasks, and billions of paired RGB and tactile frames. The corpus combines real-robot teleoperation with UMI demonstrations, jointly providing physically grounded robot executions and scalable human-operated trajectories. By recording tactile signals alongside visual observations, NeoData captures contact states, local forces, and incipient slip that are difficult to infer from vision alone. To support open research, we further release OpenNeoData, a $5{,}000$-hour open-source subset spanning six embodiments, more than $250$ tasks, and over $200$ skills.
Based on the large-scale and heterogeneous tactile data in NeoData, we introduce a unified, hardware-agnostic tactile representation learning formulation that uses dense three-axis force fields as a common physical supervision space across different tactile sensor designs. Under this formulation, we develop NeoForce, a visuo-tactile representation model that learns transferable and temporally structured tactile representations for downstream embodied policies. Finally, we provide standardized evaluation for contact-rich manipulation by combining the real-world NeoReal suite with the simulated NeoSim suite. 

In summary, this report makes the following five contributions:
\begin{itemize}
    \item \textbf{An overall tactile-centric paradigm for embodied manipulation.} We establish $\mathcal{N}_0$-Foundation, a comprehensive embodied paradigm that unifies a scalable tactile infrastructure, large-scale tactile data acquisition, hardware-agnostic tactile representation learning, and standardized policy evaluation.
    \item \textbf{A scalable tactile-based embodied infrastructure.} We develop a comprehensive data collection hardware and pipeline, including durable vision-based tactile sensors and $\mathcal{N}_0$-TacUMI, a sensor-integrated UMI device that enables efficient tactile manipulation data collection at scale.
    \item \textbf{A large-scale cross-embodiment visuo-tactile dataset.} We release \emph{NeoData}, a large-scale cross-embodiment corpus that combines real-robot teleoperation and UMI demonstrations, covers six embodiments and $450+$ tasks, and augments RGB observations with synchronized tactile sensing for contact-rich manipulation.
    \item \textbf{A unified force-based tactile representation.} We develop \emph{NeoForce}, a tactile-oriented model that unifies versatile tactile signals into a hardware-agnostic representation via dense three-axis force fields, providing a transferable interface for tactile application across embodied models.
    \item \textbf{A standardized evaluation protocol.} We comprehensively evaluate tactile-aware policies on the proposed real-world \emph{NeoReal} and simulated \emph{NeoSim} test suites, demonstrating the effectiveness of large-scale tactile learning and unified tactile representation learning.

\end{itemize}

Collectively, these contributions establish a scalable path for incorporating tactile perception into large-scale embodied learning, enabling policies to acquire fine-grained contact-rich manipulation capabilities beyond vision alone.

\section{Related Work}

\subsection{Embodiment Hardware and Data Collection}
Embodied manipulation systems couple the robot body, sensing stack, and demonstration interface. Conventional teleoperation records executable robot actions directly but binds collection to particular arms and grippers. UMI \citep{chi2024universal} instead uses a handheld gripper and relative end-effector commands to decouple demonstration collection from physical robots, and subsequent systems extend this principle to mobile, dexterous, and hardware-independent platforms \citep{ha2024umi,xu2025dexumi,liu2024fastumi}. Portable visuo-tactile interfaces (Touch in the Wild \citep{zhu2026touch}, FreeTacMan \citep{wu2025freetacman}, ViTaMIn-B \citep{li2025vitamin}, TacUMI \citep{cheng2026tacumi}, and exUMI \citep{xu2025exumi}) further attach contact sensors to handheld or wearable collectors. TAMEn \citep{wu2026tamen} additionally combines a cross-morphology wearable interface, MoCap/VR tracking, online feasibility checking, and tactile recovery teleoperation in a closed-loop collection engine. Yet these systems generally target a few robot platforms and retain device-specific tactile streams. NeoData combines robot-free collection with multi-robot teleoperation and maps six embodiments to a common action and observation schema.

\subsection{Embodied Manipulation Datasets}
Large-scale datasets such as BridgeData V2 \citep{walke2023bridgedata}, Open X-Embodiment \citep{oneill2024open}, and DROID \citep{khazatsky2024droid} establish data scale and embodiment diversity as central ingredients of general-purpose robot learning, but their observations remain dominated by vision and proprioception. RH20T is a multimodal exception, providing more than 110k real-robot sequences over 147 tasks with synchronized vision, force, audio, and action streams, although fingertip tactile signals appear in only one of its seven configurations \citep{rh20t}. 
At a substantially larger scale, AgiBot World provides more than one million teleoperated trajectories spanning 217 tasks and 2,976.4 hours. It uses over 100 homogeneous mobile dual-arm robots equipped with RGB-based visuo-tactile sensors, and applies human-in-the-loop quality verification \citep{AgiBotWorldTeam2025agibot-world-colosseo}.
The portable visuo-tactile datasets above improve contact coverage but remain at the scale of $10^{2}$--$10^{4}$ demonstrations and split across heterogeneous tactile encodings, ranging from low-resolution taxel arrays to camera-based RGB streams. Recent corpora such as XenseRobotics-TacVerse \citep{tacverse2026datasets} and Daimon-Infinity \citep{daimonrobotics_daimon_infinity_2026} substantially expand the scale and duration of robot-free visuo-tactile collection, but contain no demonstrations executed on physical robot platforms. These datasets are also typically gathered under a single acquisition mode, either teleoperation or a robot-free handheld device, and seldom document any per-episode quality control. Most closely related, OmniVTA \citep{omnivta} contains 21,879 visuo-tactile-action trajectories over 86 tasks and 126 objects and is the only prior corpus that pairs robot-based and handheld collection, though still without verified quality control. As Tab.~\ref{tab:related-work-dataset-comparison} summarizes, NeoData scales this convergence to 1.4M+ episodes over 450 tasks and more than 30,000 hours. It combines on-robot teleoperated demonstrations with scalable robot-free collection across six embodiment types, standardizes every finger on camera-based tactile RGB frames that are further converted into a unified three-axis force representation, and passes each episode through automatic and human quality control.

\begin{table}[!ht]
\centering
\footnotesize
\setlength{\tabcolsep}{3pt}
\renewcommand{\arraystretch}{1.02}
\definecolor{NeoBest}{HTML}{C00000}
\definecolor{NeoSecond}{HTML}{1F5FBF}
\begin{tabularx}{\linewidth}{@{}>{\raggedright\arraybackslash}X >{\centering\arraybackslash}p{0.11\linewidth} >{\centering\arraybackslash}p{0.085\linewidth} >{\centering\arraybackslash}p{0.13\linewidth} >{\centering\arraybackslash}p{0.10\linewidth} >{\centering\arraybackslash}p{0.11\linewidth} >{\centering\arraybackslash}p{0.115\linewidth}@{}}
\toprule
\textbf{Dataset} & \textbf{Traj.\,/\,tasks} & \textbf{Time} & \textbf{Embodiment Types} & \textbf{Tactile signal} & \textbf{Control Type} & \textbf{Verified QC} \\
\midrule
\multicolumn{7}{c}{\textbf{Large-scale robot manipulation corpora}} \\
\midrule
BridgeData V2 \citep{walke2023bridgedata} & 60k\,/\,13 & -- & 1 & -- & Tele & -- \\
Open X-Embodiment \citep{oneill2024open} & \textcolor{NeoSecond}{1M+}\,/\,-- & -- & \textcolor{NeoBest}{22} & -- & Tele & -- \\
DROID \citep{khazatsky2024droid} & 76k\,/\,86 & 350\,h & 1 & -- & Tele & -- \\
RH20T \citep{rh20t} & 110k\,/\,147 & -- & 4 & taxel & Tele & Human \\
AgiBot World \citep{AgiBotWorldTeam2025agibot-world-colosseo}
& \textcolor{NeoSecond}{1M+}\,/\,217
& \textcolor{NeoSecond}{2,976.4\,h}
& 1
& RGB
& Tele
& Human \\
\midrule
\multicolumn{7}{c}{\textbf{Robot-free collection interface}} \\
\midrule
UMI \citep{chi2024universal} & 1.4k\,/\,-- & 12\,h & 2 & -- & RF & -- \\
\midrule
\multicolumn{7}{c}{\textbf{Visuo-tactile manipulation corpora}} \\
\midrule
Touch in the Wild \citep{zhu2026touch} & 2.7k+\,/\,43 & 31\,h & 1 & taxel & RF & -- \\
FreeTacMan \citep{wu2025freetacman} & 10k+\,/\,50 & 27.8\,h & 2 & RGB & RF & -- \\
ViTaMIn-B \citep{li2025vitamin} & 0.8k\,/\,4 & 2.9\,h & 2 & RGB & RF & -- \\
TacUMI \citep{cheng2026tacumi} & 30k\,/\,1 & 0.5\,h & 1 & RGB & RF & -- \\
exUMI \citep{xu2025exumi} & 1.7k\,/\,9 & 5\,h & 1 & RGB & RF & -- \\
XenseRobotics-TacVerse \citep{tacverse2026datasets} & 8.2k\,/\,128 & 200\,h & 1 & RGB & RF & -- \\
Daimon-Infinity \citep{daimonrobotics_daimon_infinity_2026}
& 276k\,/\,-
& 1,209\,h
& 1
& RGB
& RF
& -- \\
\midrule
\textbf{NeoData (ours)} & \textcolor{NeoBest}{\textbf{1.4M+\,/\,450+}} & \textcolor{NeoBest}{\textbf{$>$30,000\,h}} & \textcolor{NeoSecond}{6} & RGB & \textbf{Tele+RF} & \textbf{Auto+Human} \\
\bottomrule
\end{tabularx}
\caption{Representative manipulation datasets in three families. Traj.\,/\,tasks: counts as stated by each paper (BridgeData V2: skills; TacUMI: frames), rounded to thousands (k). Time: total interaction time in hours, estimated from frame counts and rates where not directly reported. Embodiment Types: number of distinct embodiment platforms ($N{\times}$ identical arms count once). Tactile signal: released contact modality, grouped as camera-based RGB or taxel arrays. Control Type: collection mode, teleoperation (Tele), robot-free handheld (RF), or both (Tele+RF). Verified QC: documented per-episode quality control. ``--'': not reported or not applicable. \textcolor{NeoBest}{Red} = column best, \textcolor{NeoSecond}{blue} = second best.}
\label{tab:related-work-dataset-comparison}
\end{table}

\subsection{Robot Learning in Representation and Manipulation}
Robot learning increasingly combines scalable policy learning with reusable perception. RT-1 \citep{brohan2022rt1} and Diffusion Policy \citep{chi2023diffusion} demonstrate effective action modeling for real-world manipulation, and VLA models such as $\pi_{0.5}$ scale this recipe into generalist policies \citep{black2025pi05}. Recent embodied foundation models extend this scaling thesis. Xiaomi-Robotics-0 couples flow-matching action generation with asynchronous inference for real-time bimanual control \citep{cai2026xiaomirobotics}, Xiaomi-Robotics-1 pre-trains on more than 100k hours of real-world trajectories collected with handheld UMI devices \citep{xiaomi2026robotics1}, LingBot-VLA 2.0 curates roughly 60,000 hours of robot and egocentric human data spanning 20 robot configurations and whole-body action spaces \citep{wu2026lingbotvla2}, and the HY-Embodied series builds Mixture-of-Transformers (MoT) and Mixture-of-Experts (MoE) vision language models that support downstream VLA training \citep{tencent2026hyembodied05,wang2026hyembodiedvlm}, while concurrent systems unify understanding, generation, and action or simplify the VLA stack \citep{cai2026internvla,ye2026starvla}. A complementary line uses video and world modeling for control. LingBot-Vision pretrains dense spatial representations via masked boundary modeling \citep{fu2026lingbotvision}, LingBot-Video scales MoE video pretraining on robot-oriented footage \citep{ma2026lingbotvideo}, causal video-action models such as LingBot-VA 2.0 predict scene evolution jointly with actions \citep{li2026causal,zhang2026lingbotva2}, world-action models integrate world prediction into policy learning \citep{niu2026t, yuan2026ftp, lou2026dream}, and interactive or 3D world models serve as simulators and embodied data engines \citep{gao2026lingbotworld2,tencent2026hyworld2}, with Xiaomi-Robotics-U0 substantially improving the out-of-distribution success rate of $\pi_{0.5}$ through synthesized embodied data \citep{li2026xiaomiu0}. Across these foundation-model efforts, however, observations and pretraining corpora remain visual, linguistic, and proprioceptive, while contact is neither sensed nor standardized, and TouchGuide's tactile contact model, which steers pretrained visuomotor policies at inference time, remains a rare exception \citep{zhang2026touchguide}.

On the perception side, masked reconstruction, joint-embedding prediction, and self-supervised visual backbones provide transferable representation-learning objectives \citep{assran2023self,oquab2023dinov2}. Tactile observations are harder to reuse, because raw images from sensors such as GelSight \citep{yuan2017gelsight} and DIGIT \citep{lambeta2020digit} entangle contact with sensor-specific optics, geometry, and elastomer appearance. Prior work reduces this dependence through canonical force-based pretraining \citep{wu2025canonical}, sensor-invariant contact geometry \citep{sitr}, action-aware tactile latents \citep{xu2025exumi, xu2026seeing}, or continuous deformation representations \citep{omnivta}. NeoData instead uses a calibrated field of $x$-, $y$-, and $z$-axis forces as the dataset-level interchange format, and NeoForce learns over this shared physical interface so that heterogeneous tactile demonstrations can support a common manipulation policy.

\subsection{Embodied Manipulation Benchmarks}
Simulation suites such as RLBench \citep{james2020rlbench}, Meta-World \citep{yu2020metaworld}, CALVIN \citep{mees2022calvin}, ManiSkill2 \citep{gu2023maniskill2}, and LIBERO \citep{liu2023libero} standardize multi-task, language-conditioned, and lifelong manipulation, and household-scale successors further extend task and scene diversity \citep{nasiriany2024robocasa,nasiriany2026robocasa365,li2023behavior,zhang2025vlabench}. Perturbation studies show, however, that near-saturated scores on such suites can reflect overfitting to the specific task configurations seen during training rather than task understanding \citep{zhou2025liberopro,fei2025liberoplus}, motivating capability-level diagnosis of generalist policies \citep{gao2026ebench}. Evaluation is therefore anchoring itself to physical hardware. FurnitureBench \citep{heo2023furniturebench} standardizes long-horizon real-world assembly, SIMPLER \citep{li2024simpler} builds simulated twins whose scores predict real performance, and RoboDojo \citep{chen2026robodojo} and DuoBench \citep{julg2026duobench} couple simulated and real task suites under shared protocols, which is the paired sim-and-real methodology that we adopt. All of these benchmarks, however, observe only vision and proprioception, so contact quality is neither sensed nor scored.

Tactile manipulation benchmarks remain scarce because simulation must reproduce both contact mechanics and sensor deformation. Taxim \citep{si2022taxim} and TacSL \citep{akinola2024tacsl} simulate the appearance of specific vision-based tactile sensors, UniVTAC \citep{chen2026univtac} unifies multi-sensor tactile simulation with an eight-task benchmark, ContactWorld \citep{contactworld} evaluates visuo-tactile world models on twelve simulated tasks and one real task, and SoftVTBench \citep{jing2026softvtbench} finds that policies often succeed while violating deformation limits, a failure that integrating tactile sensing substantially reduces. These efforts substantially expand tactile evaluation within their own scope, but their data remains tied to a particular sensor or confined to simulation, so the resulting protocols do not transfer across tactile hardware or to physical deployment. In our work, both NeoReal and NeoSim are paired for policy evaluation, so that tactile feedback and representation choices can be compared reproducibly without tying evaluation to the appearance of any particular sensor.

\FloatBarrier

\begin{figure}[t]
    \centering
    \includegraphics[width=\linewidth]{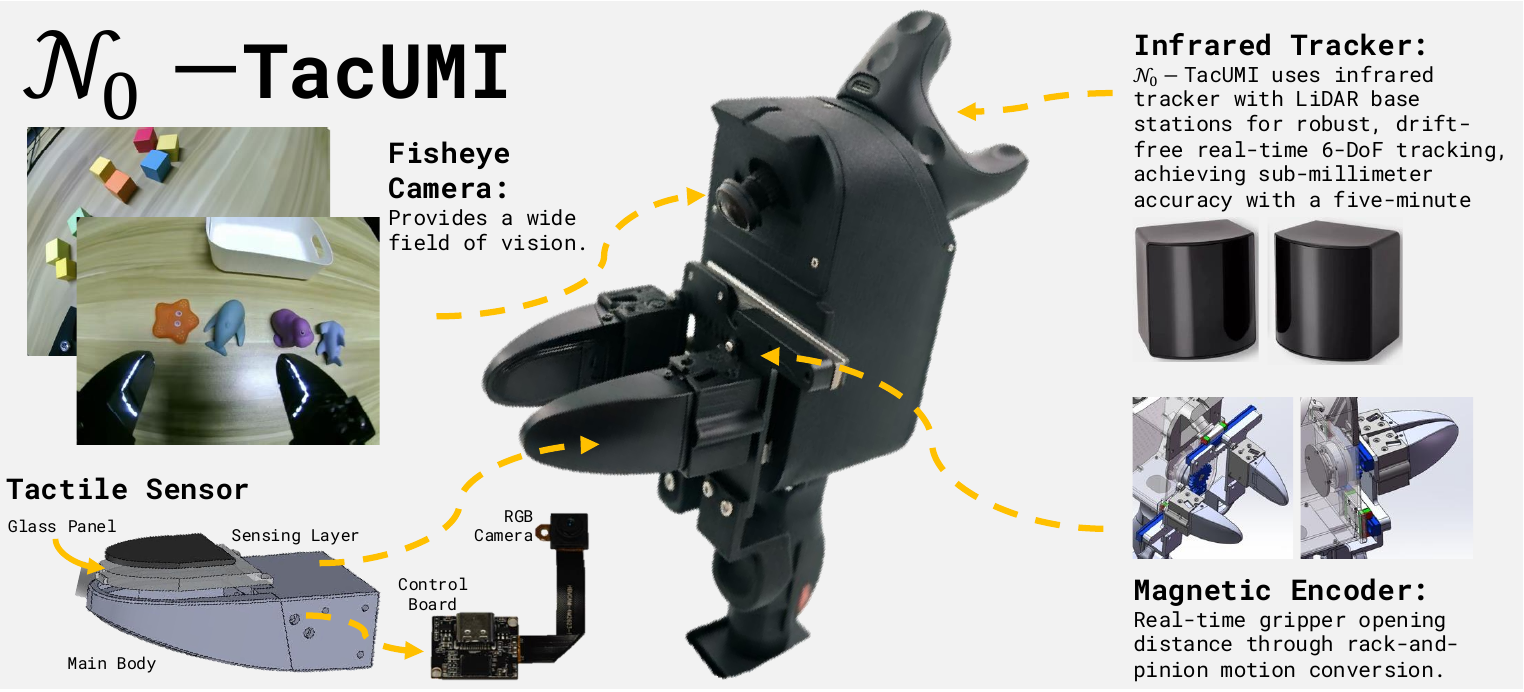}
    \caption{Overview of $\mathcal{N}_0$-TacUMI. A single handheld unit records complementary streams: a fisheye camera captures the wide-angle wrist view, two camera-based tactile sensors capture contact-rich deformation at the fingertips, an infrared tracker with base stations provides a low-drift six-degree-of-freedom pose, and a magnetic encoder measures the gripper aperture through a rack-and-pinion transmission. The inset shows the layered structure of a tactile sensor, comprising a glass panel, an elastic sensing layer, the main body with an embedded RGB camera, and a controller board.}
    \label{fig:neote-umi}
\end{figure}

\section{Data Acquisition Hardware}

\subsection{Overview}
NeoData spans six embodiments, comprising five robot platforms (Franka Research 3, Piper, ARX X5, UR5e, and Flexiv Rizon 4s robots) and our handheld $\mathcal{N}_0$-TacUMI device. To collect comprehensive, high-quality tactile demonstrations, we adopt two complementary acquisition settings. For the five robot embodiments, tactile sensors are mounted directly on the physical grippers, and demonstrations are collected through teleoperation on the corresponding target platforms. In parallel, $\mathcal{N}_0$-TacUMI is a human-operated handheld parallel gripper developed following the Universal Manipulation Interface philosophy \citep{chi2024universal}, enabling scalable demonstration collection beyond fixed robot workcells. All six embodiments share aligned tactile sensing, visual observations, and action conventions, allowing NeoData to integrate physically grounded robot executions and scalable TacUMI demonstrations within a unified data schema.

\subsection{Tactile Sensor}
\label{subsec:tactile-sensor}
Our tactile sensor follows the camera-based visuo-tactile principle \citep{yuan2017gelsight, lambeta2020digit, luo2024novel, luo2025efficacy, luo2025quantitative}, in which contact information is recovered from the optical distortion of a deformable sensing surface. As illustrated in the inset of Fig.~\ref{fig:neote-umi}, the sensor is a layered assembly. A thin wear-resistant glass panel protects the contact surface against abrasion during repeated interaction. Beneath it, an elastic sensing layer deforms under external load and carries a textured pattern whose displacement encodes the applied force. An RGB camera embedded in the main body observes the underside of the sensing layer, while a compact controller board handles illumination, image acquisition, and data transmission.

When the finger contacts an object, local pressure and shear deform the sensing layer, and the camera captures the resulting texture displacement as a tactile image $T_t \in \mathbb{R}^{H \times W \times 3}$. The tactile signal is therefore a visual measurement of deformation. Let $\Phi$ denote the elastomer force-to-deformation transfer function, which maps a spatial force distribution $F$ to a deformation field, and let $\mathcal{C}$ denote the imaging process of the internal camera. The observed tactile image can be written as
\begin{equation}
    T_t = \mathcal{C}\big( \Phi( F_t ) \big),
\end{equation}
so that recovering the underlying force field $F_t$ from $T_t$ amounts to inverting the composed transduction and imaging operators. We learn this inverse mapping explicitly and use its output as the unified tactile representation, as detailed in Section~\ref{sec:tactile-representation}.

\subsection{$\mathcal{N}_0$-TacUMI}
\label{subsec:tactile-umi}
Building on the tactile sensor, we develop $\mathcal{N}_0$-TacUMI for robot-free data collection, shown in Fig.~\ref{fig:neote-umi}. The device consists of a parallel gripper equipped with two tactile sensors, a $160^\circ$ FOV fisheye camera that captures the wide-angle wrist view, an infrared tracker for spatial localization, and a magnetic encoder for physical aperture measurement. The tracker provides the six-degree-of-freedom gripper pose in a global tracking frame, while the magnetic encoder reads the instantaneous finger separation through a rack-and-pinion transmission.

$\mathcal{N}_0$-TacUMI records synchronized wrist images, left and right tactile images, gripper aperture, and relative end-effector motion. Following the UMI convention \citep{chi2024universal,kroger2011manipulation}, actions are represented by the end-effector pose and the gripper width. This convention avoids binding the demonstration to a specific robot base frame and makes the handheld trajectories compatible with the teleoperated robot data described in Section~\ref{sec:neodata}. The detailed observation and action definitions are provided in Section~\ref{subsec:data-format}.

\begin{figure}[t]
    \centering
    \includegraphics[width=0.98\linewidth]{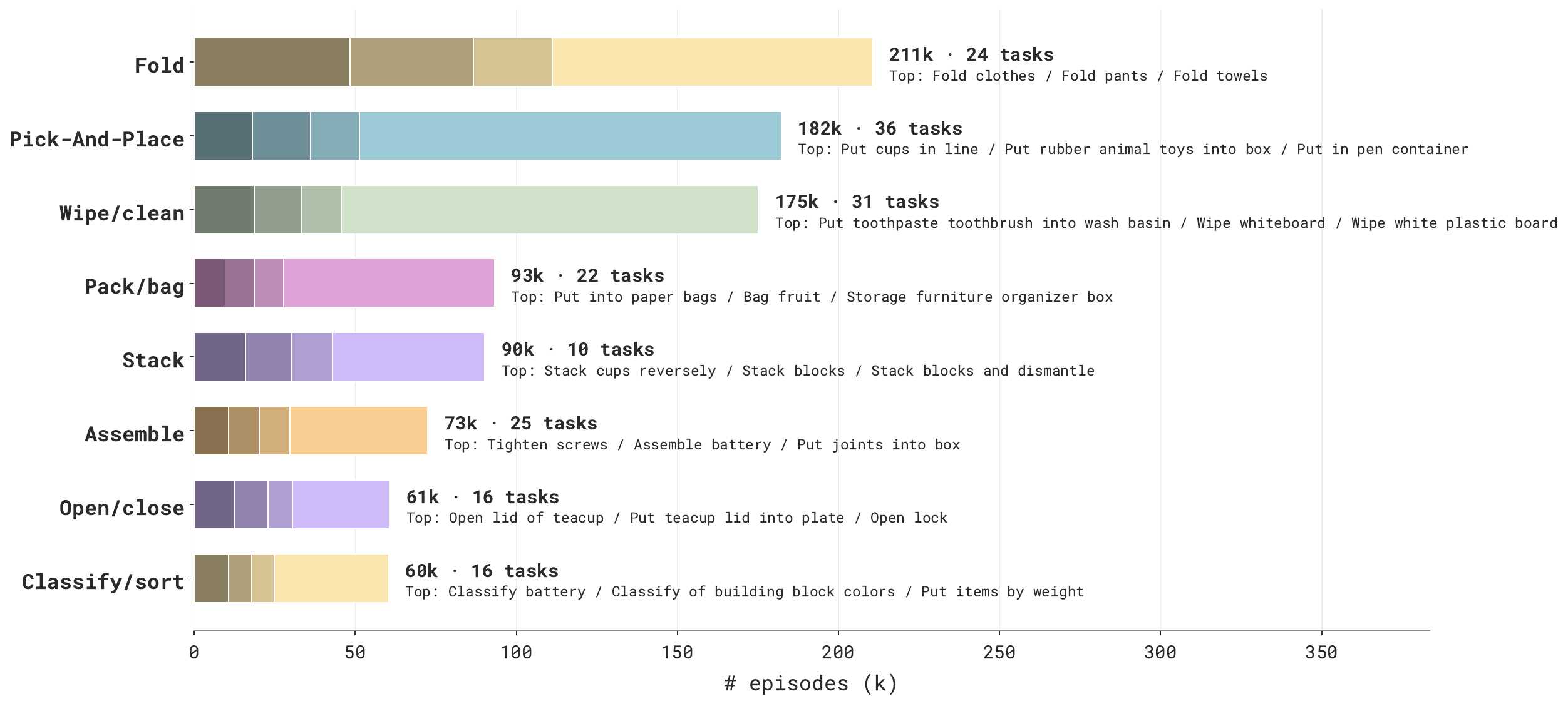}
    \caption{Task coverage of NeoData. The more than $450$ tasks are organized into eight categories, and for each category the most frequent constituent tasks are shown by episode count. The categories span deformable-object skills, high-precision assembly, everyday transport, surface interaction, and semantic manipulation.}
    \label{fig:neodata-task-classification-bar}
\end{figure}

\section{NeoData}
\label{sec:neodata}

\subsection{Overview}

We introduce NeoData, one of the largest tactile-enabled real-world manipulation datasets constructed to date. As summarized in Fig.~\ref{fig:teaser}, the corpus aggregates more than $30{,}000$ hours of interaction into $1.4$M episodes and $3.3$B timesteps, yielding $8$B RGB frames paired with $10$B tactile frames. The data span six embodiments, including the handheld $\mathcal{N}_0$-TacUMI device and five robot platforms: ARX X5, UR5e, Flexiv Rizon 4s, Piper and Franka Research 3 robots, and were collected by nearly 100 human operators. Every episode contains synchronized visual and tactile streams, so the contact signal is available throughout the manipulation process rather than only at discrete grasp events. To facilitate open research, we release OpenNeoData, a $5{,}000$-hour open-source subset of NeoData that spans six embodiments, more than $250$ tasks, and over $200$ skills, providing an accessible entry point to the full corpus.

Tab.~\ref{tab:related-work-dataset-comparison} situates NeoData among representative manipulation datasets. Compared with large visual corpora, NeoData adds dense tactile observations at comparable scale and covers a broad set of contact-rich skills. Compared with existing tactile datasets and collection systems, it provides larger episode volume, greater embodiment diversity, and a mixture of real-robot and UMI demonstrations. The dataset is therefore designed to support policy learning in settings where tactile feedback is not incidental but central to the manipulation objective.

\subsection{Data Format}
\label{subsec:data-format}

Every NeoData episode is stored as a synchronized sequence of observations and actions. For the five robot embodiments, the observation at timestep $t$ contains an external third-person RGB image $I_t^{\mathrm{ext}}$, a wrist RGB image $I_t^{\mathrm{wrist}}$, and the left and right tactile images, denoted as $T_t^{l}$ and $T_t^{r}$, respectively:
\begin{equation}
    o_t^{\mathrm{robot}} =
    \left( I_t^{\mathrm{ext}},\; I_t^{\mathrm{wrist}},\; T_t^{l},\; T_t^{r} \right),
\end{equation}
where each image stream is temporally aligned and the tactile images are recorded at a native resolution of $640 \times 360$. The robot action $a_t^{\mathrm{robot}}$ contains the end-effector command in the tool center point frame and the absolute joint command:
\begin{equation}
    a_t^{\mathrm{robot}} =
    \left( \Delta x_t^{\mathrm{tcp}},\; q_{t+1} \right),
\end{equation}
where $\Delta x_t^{\mathrm{tcp}}$ denotes the commanded Cartesian end-effector motion and $q_{t+1}$ denotes the target joint configuration for the next control step. We represent rotations using the continuous 6D representation to avoid the discontinuities of Euler-angle and quaternion parameterizations, as well as the gimbal-lock issue associated with Euler angles. This format preserves both the operational action used by task policies and the absolute robot state needed for embodiment-specific replay or analysis.

For $\mathcal{N}_0$-TacUMI data, the observation $o_t^{\mathrm{UMI}}$ contains a fisheye wrist image $I_t^{\mathrm{fisheye}}$ and the left and right tactile images:
\begin{equation}
    o_t^{\mathrm{UMI}} =
    \left( I_t^{\mathrm{fisheye}},\; T_t^{l},\; T_t^{r} \right).
\end{equation}
The corresponding action records the relative end-effector motion between the current frame and a future frame, together with the target gripper aperture:
\begin{equation}
    a_t^{\mathrm{UMI}} =
    \left( \Delta x_{t \rightarrow t+k},\; g_{t+k} \right),
\end{equation}
where $\Delta x_{t \rightarrow t+k}$ denotes the relative motion between frame $t$ and $t + k$, and $g_{t+k}$ denotes the gripper aperture at frame $t + k$.
This relative action convention follows UMI and supports transfer from handheld demonstrations to robot embodiments.

\begin{figure}[t]
    \centering
    \includegraphics[width=0.98\linewidth]{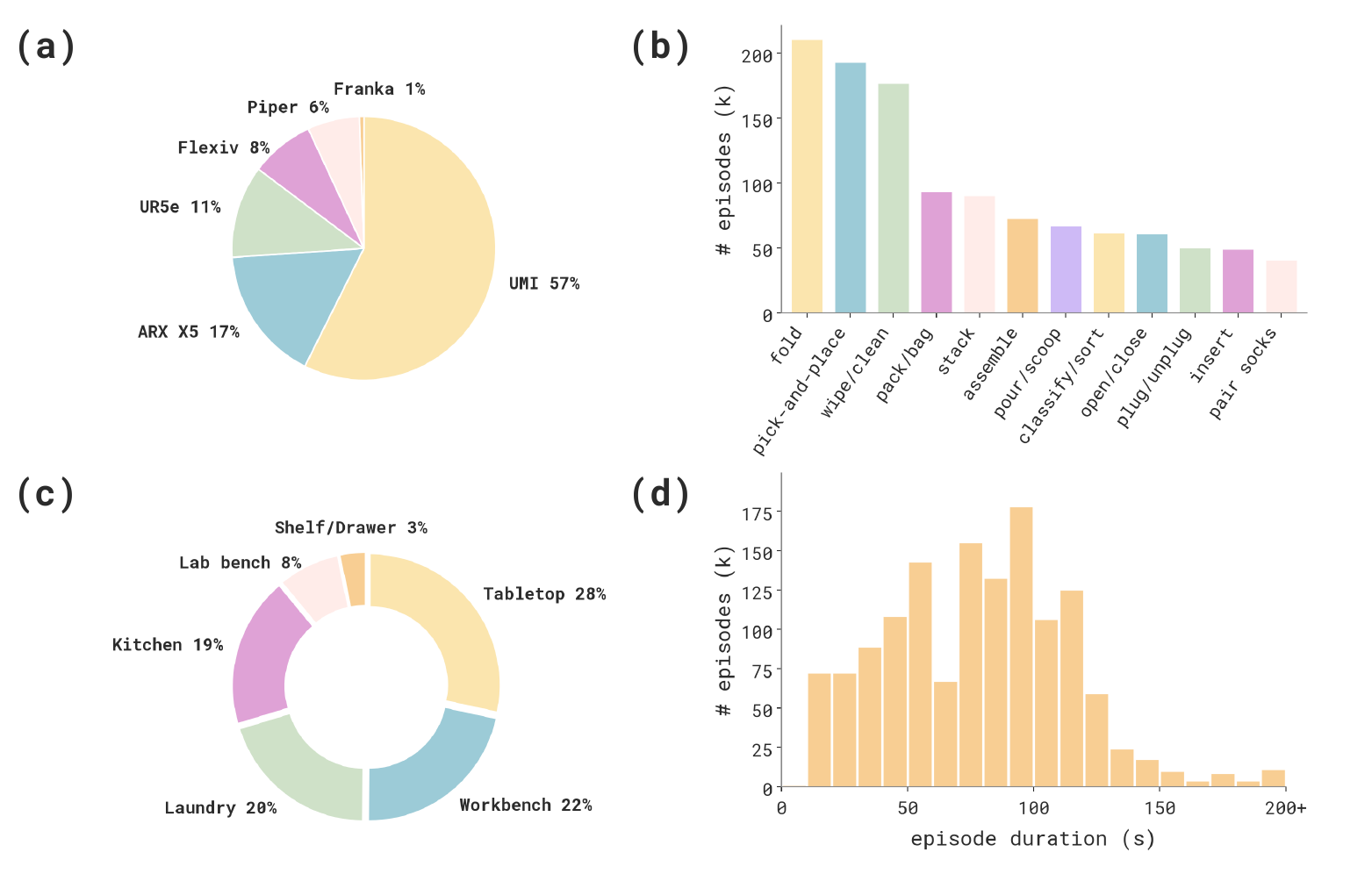}
    \caption{NeoData statistics. (a) Trajectories per embodiment: $\mathcal{N}_0$-TacUMI contributes $57\%$ of the data, followed by ARX X5 ($17\%$), UR5e ($11\%$), Flexiv Rizon 4s ($8\%$), Piper ($6\%$), and Franka Research 3 ($1\%$). (b) Episode counts for the twelve most frequent object categories, led by cups ($33.8$k), boxes ($32.6$k), test tubes ($22.3$k), and building blocks ($21.1$k). (c) Scene distribution across tabletop, workbench, laundry, kitchen, lab bench, and shelf or drawer settings. (d) Episode duration histogram, which peaks near $80$ seconds with a long right tail.}
    \label{fig:neodata-statistics}
\end{figure}

\subsection{Dataset Composition}

Fig.~\ref{fig:neodata-statistics} summarizes the embodiment, object, scene, and duration distributions, while Fig.~\ref{fig:neodata-task-classification-bar} summarizes task coverage. $\mathcal{N}_0$-TacUMI contributes $57\%$ of all trajectories, reflecting the scalability of handheld collection for dexterous and repetitive tasks. The five robot embodiments provide the remaining trajectories and anchor the corpus in executable robot behavior. This mixture allows NeoData to combine the efficiency of UMI demonstrations with the embodiment fidelity of robot teleoperation. The object distribution is likewise long-tailed, with cups and boxes as the most frequent object categories, followed by test tubes and building blocks.

The task distribution deliberately emphasizes tactile-relevant manipulation. High volume categories include folding, pick-and-place, wiping or cleaning, stacking, assembling, inserting, pouring, opening, closing, and classification. These categories cover deformable-object manipulation, precise mating, delicate grasping, sustained surface contact, and manipulation under ambiguous visual states. Scenes include tabletop, workbench, laundry, kitchen, lab bench, and storage environments, with objects spanning containers, tools, deformables, kitchenware, electronics, and lab items. Episode durations peak near $80$ seconds and include a long tail of multistep interactions, which supports learning over both short contact primitives and extended task sequences.

\begin{figure}[t]
    \centering
    \includegraphics[width=0.98\linewidth]{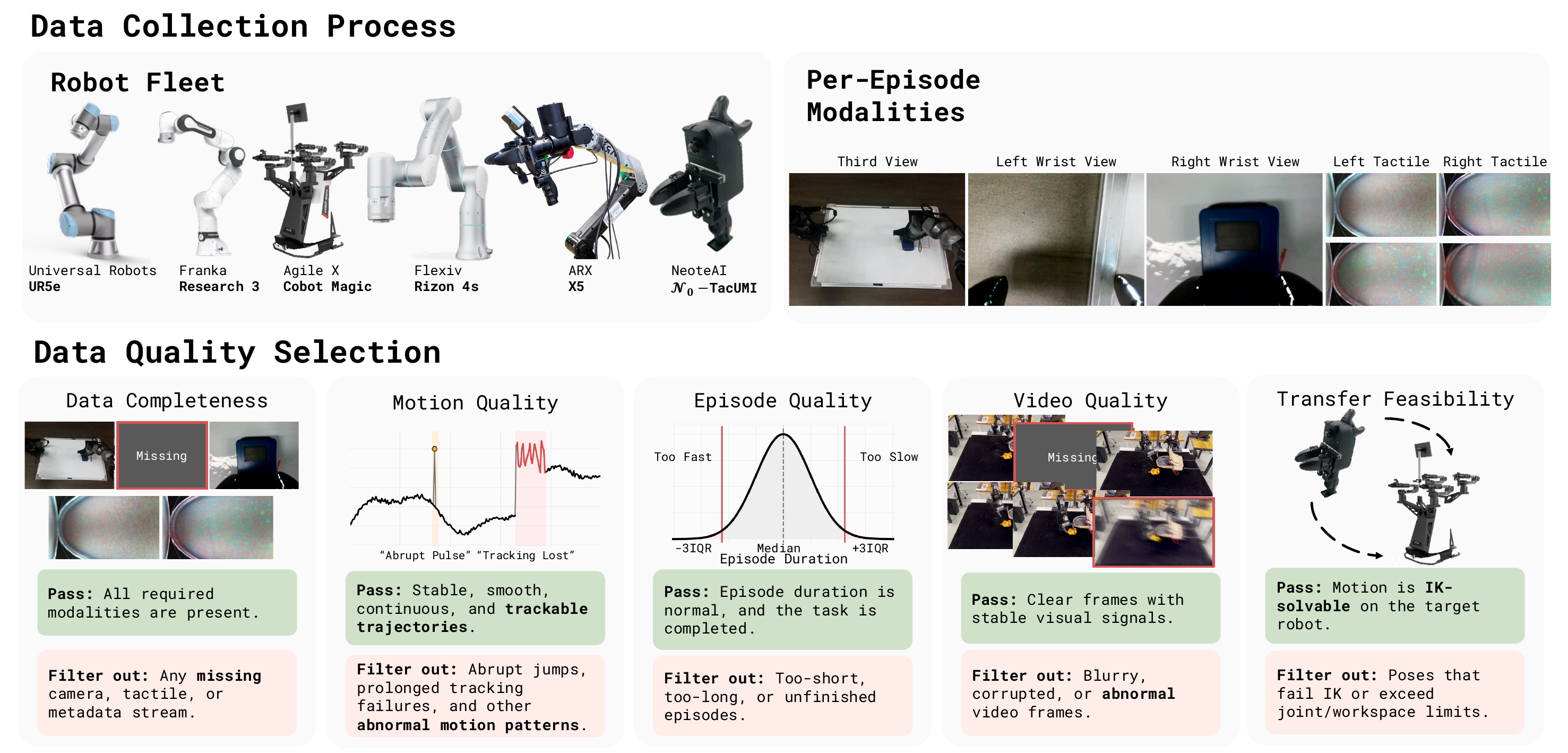}
    \caption{The NeoData curation pipeline. Raw demonstrations are stored as synchronized multimodal episodes and pass through quality checks for stream completeness, motion quality, episode quality, video validity, and transfer feasibility before entering the released corpus.}
    \label{fig:data-curation-pipeline}
\end{figure}

\subsection{Data Curation and Preprocessing}

Fusing real-robot and $\mathcal{N}_0$-TacUMI demonstrations requires effectively integrating cross-embodiment data and unifying it into a consistent format. This motivates a multi-step cleaning pipeline that turns heterogeneous raw recordings into a complete, high-quality corpus suitable for model training. As illustrated in Fig.~\ref{fig:data-curation-pipeline}, we summarize this process into a staged quality selection pipeline of five complementary checks.

\paragraph{Data Completeness.}
The first check verifies that each episode contains the required visual, tactile, action, and metadata streams. Episodes that are missing a required camera, tactile, action, or metadata stream are removed so that each retained sequence provides a complete multimodal record.

\paragraph{Motion Quality.}
The second check screens for motion artifacts that would destabilize policy training. We estimate trajectory smoothness along each episode and flag low-quality regions such as abrupt tracking pulses and tracking-lost segments, and episodes dominated by these artifacts are removed.

\paragraph{Episode Quality.}
The third check normalizes the pace of the demonstrations and verifies that each one actually completes its task. For each task, we compute the median episode duration $m$ and the difference between the 75th- and 25th-percentile durations, denoted as $r=q_{75}-q_{25}$. We retain only episodes whose duration falls within $[m-3r,\,m+3r]$. This removes demonstrations that are completed abnormally fast or slow relative to the typical duration range of the corresponding task. Episodes that terminate before the task is completed are also removed.

\paragraph{Video Quality.}
The fourth check validates the recorded streams and repairs the defects that remain recoverable. We inspect every view for corrupted content, severe blur, dropped frames, and abnormal illumination. In addition, we align the asynchronously recorded modalities to a unified timeline with a strict one to one correspondence between video, tactile and action frames. Recording errors such as swapped camera mappings, mislabeled left and right arms, and abnormal end-effector poses are repaired when possible, for example by remapping the cameras or recovering the pose through forward kinematics, and are otherwise discarded. We finally trim near-static segments, whose nearly identical observations correspond to inconsistent actions and cause the model to pause unnecessarily at inference time.

\paragraph{Transfer Feasibility.}
A large fraction of NeoData is collected robot-free, so the recorded motions are not guaranteed to be physically realizable on the target robots. The final check therefore runs inverse kinematics on each trajectory and rejects poses that fail to solve or that exceed the joint and workspace limits of the intended embodiment, which preserves the scalability of $\mathcal{N}_0$-TacUMI collection while keeping every retained demonstration executable after transfer. Episodes that pass all five checks constitute the released corpus, complete in modality, well conditioned for training, visually valid, and executable on the target embodiment.

\begin{figure}[t]
    \centering
    \includegraphics[width=0.98\linewidth]{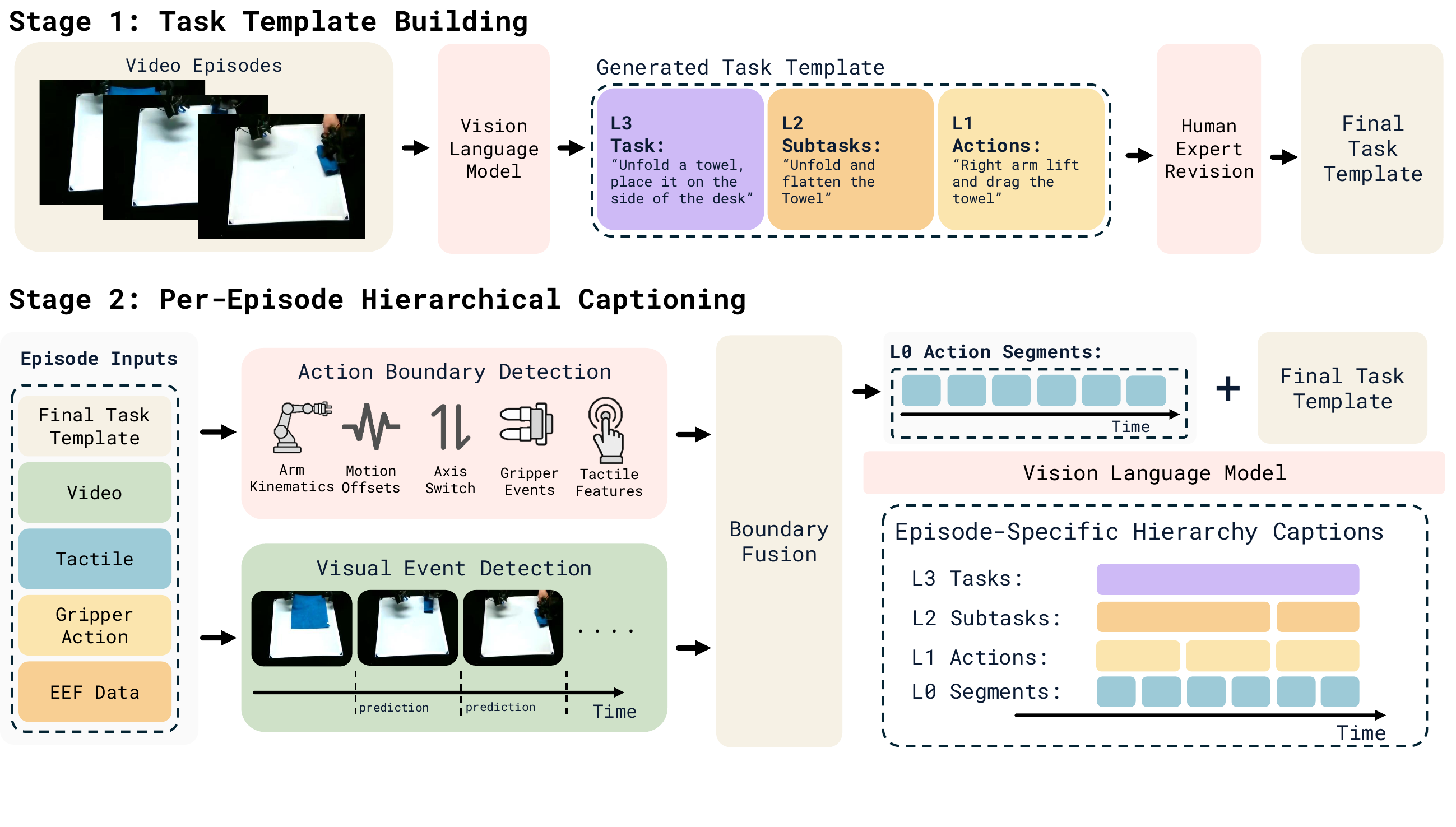}
    \caption{The NeoData hierarchical annotation pipeline. A VLM first proposes a task template from representative episodes, a human expert verifies the template, and each episode is then segmented by signal-based boundaries before the fixed intervals are labeled into a four-level hierarchy.}
    \label{fig:data-annotation-pipeline}
\end{figure}

\subsection{Data Annotation}

We introduce an automatic annotation pipeline that assigns every manipulation episode a four-level hierarchy. The levels are task (L3), subtask (L2), action (L1), and atomic segment (L0), as shown in Fig.~\ref{fig:data-annotation-pipeline}. The hierarchy separates semantic structure from temporal localization. The template defines what should occur in a task, while signal-based segmentation determines when each event occurs in a specific episode.

The pipeline has two stages. In the first stage, a vision-language model (VLM) summarizes representative episodes into a time-independent task template, which records the task objective, ordered semantic phases, and action-level substeps. A human expert verifies the template to remove semantic ambiguity before it is applied to the full task category. In the second stage, each episode is segmented by fusing action-based and visual event boundaries. The model then labels the resulting fixed intervals according to the verified template, and deterministic code assembles the L0, L1, L2, and L3 temporal extents. This design keeps temporal endpoints grounded in measured signals while using language models for semantic labeling, which reduces temporal drift on long manipulation videos. By segmenting each episode into labeled subtasks, this pipeline supplies the temporal structure needed for training long-horizon, multi-stage, and multi-task embodied policies.

\section{Tactile Representation}
\label{sec:tactile-representation}

\begin{wrapfigure}{r}{0.6\linewidth}
    \centering
    \vspace{-6pt}
    \includegraphics[width=\linewidth]{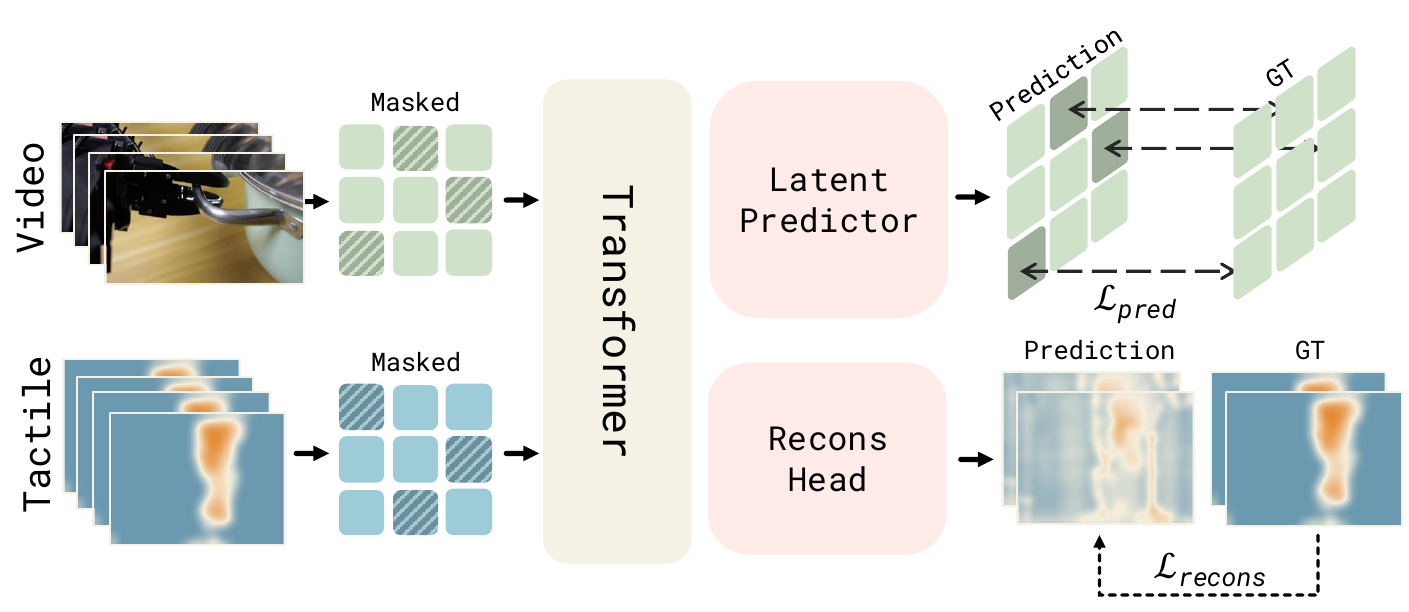}
    \caption{The structure of NeoForce. A chunk of RGB frames and tactile force fields is patchified into visual and tactile tokens and processed by a shared transformer backbone. A reconstruction head decodes force fields, while a latent prediction head predicts teacher latent outputs that serve as ground truth in the latent space.}
    \label{fig:tactile-repre-structure}
    \vspace{-10pt}
\end{wrapfigure}

This section proposes a tactile representation that allows tactile information to act on embodied manipulation policies, describing every tactile observation as a dense three-axis force field over the sensing surface and thereby expressing contact as a physical quantity rather than as the appearance of a particular device. We first present the motivation and the design principles behind the representation, then define it, describe the data pipeline that produces force fields, and present NeoForce, the model used to learn temporally structured tactile features from these force fields.

\subsection{Motivation}
\label{sec:tactile-motivation}

Unlike visual hardware, which records a consistent image space based on electronic pixels, tactile hardware is fragmented in the form of the signal it emits. Optical gel sensors~\citep{yuan2017gelsight,lambeta2020digit}, capacitive arrays~\citep{schmitz2011methods,maiolino2013flexible}, and piezoresistive skins~\citep{stassi2014flexible} differ in units, resolution, and sensing geometry, so downstream representations and models built on any single signal form are bound to the device that produced it and cannot be carried elsewhere. Scaling tactile manipulation therefore requires converting these heterogeneous signals into one universal form that the tactile community can share. To provide a transferable interface, a universal tactile representation should satisfy three principles.

\paragraph{Physical grounding.} The representation should be anchored to what touch physically delivers rather than to an appearance-level proxy. Across various manipulation scenarios, what an end-effector delivers to an object is completely described by its physical effect at the contact interface, so its behavior is expressed by an intrinsic attribute rather than by how the contact is measured through the hardware.

\paragraph{Cross-sensor unification.} The representation should be a common space that different tactile devices can utilize. Signals from any sensor with an appropriate calibration procedure should be mappable into the same target space, which makes heterogeneous tactile data directly comparable and jointly trainable.

\paragraph{Spatially rich semantics.} Complex contact is spatially structured. The location, shape, and extent of the contact patch, together with how intensity varies inside it, distinguish a full grasp from an edge contact and a stable hold from incipient slip, so the representation must preserve this distribution rather than summarize it.

\subsection{Definition}

Following the three principles above, we adopt force as the physical attribute that the tactile representation encodes, since it describes the effect of contact independently of the sensor that measured it. At every point of the sensing surface we record the local force as two orthogonal in-plane components tangent to the tactile plane, which capture the shear that drives sliding and grasping, together with one normal component, which captures the pressure exerted during pressing. Sampling this three-dimensional force on a high-resolution lattice yields a dense force field that expresses the complex contact taking place over the sensing surface and retains the spatial distribution of contact intensity. Concretely, for one tactile sensor at time $t$ the force field is
\begin{equation}
    F_t \in \mathbb{R}^{H \times W \times 3},
    \qquad
    F_t(u,v) = \big( f_x(u,v),\, f_y(u,v),\, f_z(u,v) \big),
\end{equation}
where each spatial location $(u,v)$ carries two tangential components $f_x, f_y$ along the $x$ and $y$ axes that encode shear, and one normal component $f_z$ along the $z$ axis that encodes pressure. For a parallel gripper, the left and right tactile fields are represented as $F_t^{l}$ and $F_t^{r}$ and may be concatenated into a six-channel tensor. This definition gives downstream embodied models a physically interpretable tactile input that is independent of raw gel appearance.

\subsection{Implementation}
\label{sec:tactile-pipeline}

The force representation is produced directly from raw visuo-tactile images. Let $T_t$ denote the raw tactile image from one sensor. We learn a dense mapping
\begin{equation}
    \hat{F}_t = g_\theta(T_t),
    \qquad
    g_\theta : \mathbb{R}^{H \times W \times 3} \rightarrow
    \mathbb{R}^{H \times W \times 3},
\end{equation}
where $\hat{F}_t$ is the estimated force field. The model is trained on paired visuo-tactile data, and the construction of this dataset is described in the supplementary material.

The resulting force field serves as the tactile representation of the sensor, so that given a raw tactile image, the model outputs a unified three-axis force representation of the contact. This representation also provides the target that NeoForce builds upon, since the model encodes the low-level image observation of the current sensor into a force field. Detailed calibration hardware, data split, model architecture, and preprocessing parameters are provided in Appendix~\ref{sec:supp-force-pipeline}.

\subsection{NeoForce: A Unified Force Tactile Representation Model}

NeoForce learns a temporally structured visuo-tactile representation on top of the force fields, which serves as a structured tactile representation that downstream embodied manipulation models can consume. As shown in Fig.~\ref{fig:tactile-repre-structure}, the model receives synchronized chunks of RGB observations and tactile force fields. Each modality is embedded into patch tokens, the tokens are fused by a shared transformer backbone, and temporal context is aggregated across the input chunk. This design allows the representation to encode not only the current contact state but also the recent evolution of contact. The model is supervised jointly by a reconstruction objective in the force space and a latent-space objective, so that it learns to recover physically meaningful force fields while also capturing temporally structured tactile features.

The reconstruction head predicts the force field and contact mask from the fused tokens. Let $\hat{F}$ and $\hat{M}$ denote the predicted force field and contact mask, and let $F$ and $M$ denote the corresponding targets. The reconstruction loss is
\begin{equation}
    \mathcal{L}_{\mathrm{recons}} =
    \lambda_f \left\|(\hat{F} - F) \odot M\right\|_{\mathrm{H}}
    + \lambda_g \left\|\bar{\hat{F}} - \bar{F}\right\|_{\mathrm{H}}
    + \lambda_m \mathcal{L}_{\mathrm{mask}}(\hat{M}, M),
\end{equation}
where $\|\cdot\|_{\mathrm{H}}$ is the Huber norm, $\bar{\cdot}$ denotes the spatial mean, and $\mathcal{L}_{\mathrm{mask}}$ supervises contact segmentation.

The latent prediction head supervises the model in a shared latent space to capture the temporal correlations of tactile signals across the input chunk, using a teacher model to provide the ground truth in that latent space. The teacher receives the complete input and produces latent targets, while the student receives masked inputs and predicts those targets. Let $Z_\tau$ denote the teacher latent output and $\hat{Z}_s$ denote the student prediction. The latent prediction loss is
\begin{equation}
    \mathcal{L}_{\mathrm{pred}} =
    H\!\left(Z_\tau^{v}, \hat{Z}_s^{v}\right)
    + H\!\left(Z_\tau^{t}, \hat{Z}_s^{t}\right)
    + \lambda_x H\!\left(Z_\tau^{v \leftrightarrow t}, \hat{Z}_s^{v \leftrightarrow t}\right)
    + \lambda_p H\!\left(Z_{\tau,\mathrm{patch}}, \hat{Z}_{s,\mathrm{patch}}\right),
\end{equation}
where $Z^{v}$ and $Z^{t}$ denote the visual and tactile latents. The cross-modal latent $Z^{v \leftrightarrow t}$ is obtained by predicting the tactile latent from the visual context and the visual latent from the tactile context, so that this term aligns the two modalities by forcing each to be recoverable from the other. The patch term $Z_{\mathrm{patch}}$ supervises masked patch-level latents and encourages local contact prediction under masking.

The complete objective is
\begin{equation}
    \mathcal{L} =
    \mathcal{L}_{\mathrm{recons}}
    + \mathcal{L}_{\mathrm{pred}}
    + \lambda_c \mathcal{L}_{\mathrm{con}},
\end{equation}
where $\mathcal{L}_{\mathrm{con}}$ is a contrastive alignment term between visual and tactile embeddings. We evaluate the reconstruction quality of NeoForce in Section~\ref{sec:recon}, and study how its force representation should be consumed by downstream policies in Section~\ref{sec:eval-neoreal}.

\subsection{Evaluations on Tactile Representation}
\label{sec:recon}

\begin{table}[t]
\centering
\small
\setlength{\tabcolsep}{8pt}
\renewcommand{\arraystretch}{1.15}
\begin{tabular}{l c c c}
\toprule
Objective & MAE $\downarrow$ & RMSE $\downarrow$ & mIoU $\uparrow$ \\
\midrule
Reconstruction & 0.070 & 0.095 & \textbf{0.968} \\
+ latent pred & \textbf{0.066} & \textbf{0.089} & 0.966 \\
\bottomrule
\end{tabular}
\caption{Ablation of the NeoForce training objective on force field reconstruction. Adding latent prediction preserves reconstruction fidelity while training the representation to model temporal contact evolution.}
\label{tab:recon}
\end{table}

\begin{figure}[t]
    \centering
    \includegraphics[width=0.98\linewidth]{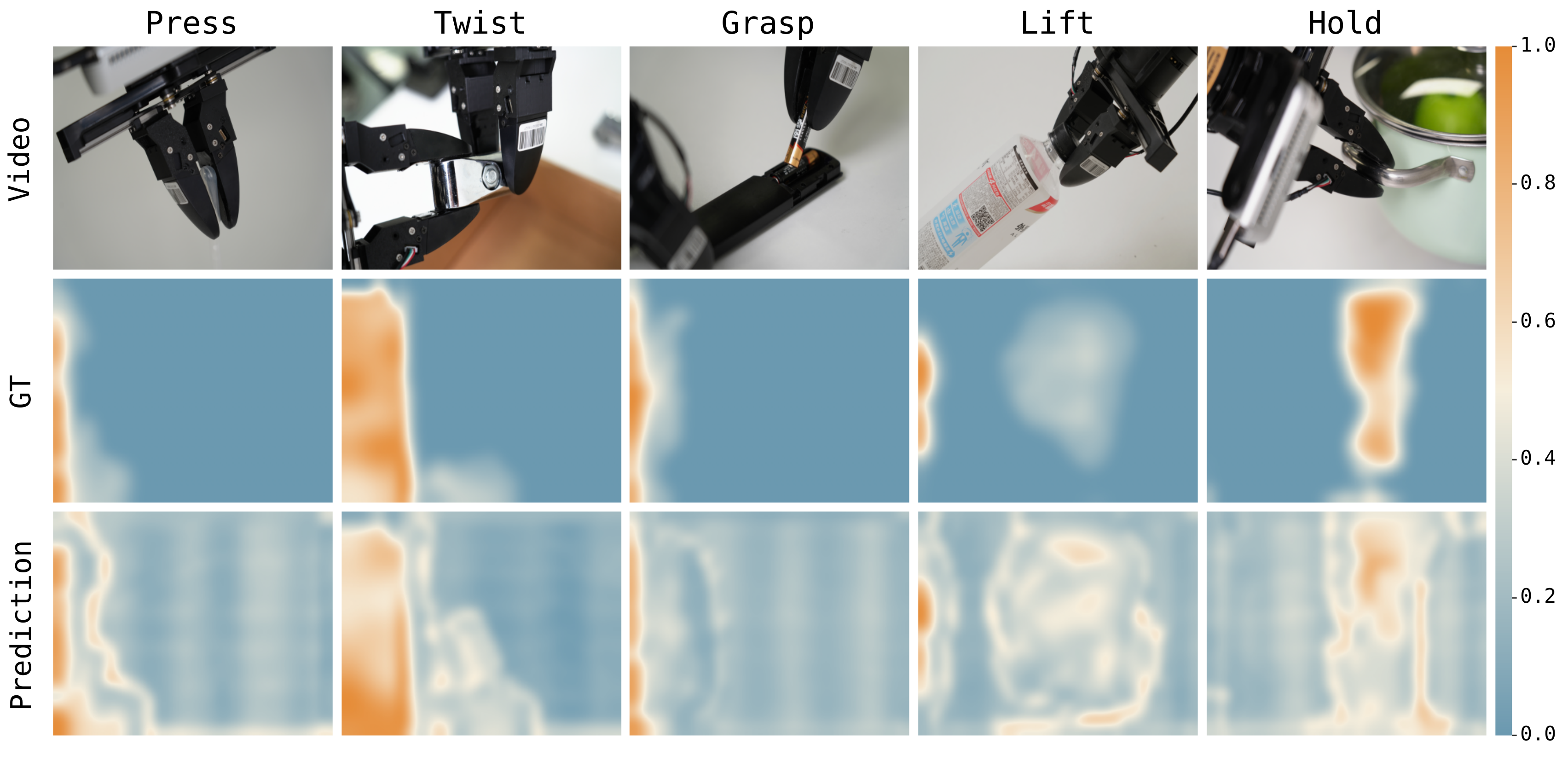}
    \caption{Force fields reconstruction of NeoForce on five representative contact behaviors: press, twist, grasp, lift, and hold. For each column, the top row shows the RGB frame, the middle row shows the ground truth force fields produced by the preprocessing pipeline of this section, and the bottom row shows the field reconstructed by NeoForce. The colorbar encodes normalized force magnitude.}
    \label{fig:tactile-repre-demonstrations}
\end{figure}

We evaluate NeoForce to figure out two essential problems, namely the effectiveness of the force-based representation on encoding contact intensity over space and time, and whether the distribution of the force space it learns stays consistent once the manipulation task varies. We first ablate the latent prediction objective against a reconstruction-only baseline and score held-out contacts with mean absolute error (MAE) and root mean squared error (RMSE) on the reconstructed force, together with mean intersection over union (mIoU) on the predicted contact region, so that errors in force magnitude and errors in contact localization are read separately. We then group the reconstructions by skill type and compare them with the ground truth force fields produced by the pipeline of Section~\ref{sec:tactile-pipeline}.

\paragraph{Finding 1. Latent prediction captures the spatial correlation of tactile signals and thereby improves the representation space.} As reported in Tab.~\ref{tab:recon}, the reconstruction-only objective already reaches an MAE of $0.070$ and an RMSE of $0.095$ together with a contact region overlap of $0.968$ mIoU. Adding latent prediction lowers the MAE to $0.066$ and the RMSE to $0.089$, while the overlap stays essentially unchanged at $0.966$ mIoU. Based on this contrast between the two metric groups, the gain is concentrated in the magnitude of the reconstructed force rather than in the localization of the contact region. Compared with the pure reconstruction objective, predicting masked patch-level and cross-modal latents requires the model to infer the contact at an unobserved location. The surrounding contact area, the neighboring frames of the chunk, and the visual context covering the same region jointly constrain that missing location, which drives the representation to convey how contact intensity is distributed across space and time. The large-scale training on NeoData also contributes to the estimate, since it endows the shared backbone with a strong visual prior over object geometry and material. Cross-modal prediction carries this prior into the tactile tokens, so the tactile branch knows the contact region and estimates its intensity quantitatively more precisely. We therefore adopt the combined objective, whose downstream benefit we confirm on NeoReal in Section~\ref{sec:eval-neoreal}. \emph{In short, latent prediction improves force estimation at no cost to contact localization, because a visual prior learned at scale helps the representation encode contact intensity quantitatively across space and time.}

\paragraph{Finding 2. Grounding on force yields a tactile representation that stays consistent across tasks.} Fig.~\ref{fig:tactile-repre-demonstrations} visualizes the effect of the representation across skills. Across press, wrist, grasp, lift, and hold, which differ substantially in contact geometry and duration, NeoForce focuses on the region where the interaction takes place and extracts the properties that characterize it, including the intensity of the applied force, the geometry of the contact patch, and its spatial extent. In detail, press, wrist, and grasp yield localized peaks with sharp boundaries and near-zero surroundings, while lift and hold yield broader sustained distributions, and the reconstructed field consistently tracks the fine-grained intensity of the ground truth over a wide dynamic range. This consistency follows from the choice of force as the representation space. Such representation places the contact effects of the individual skills, together with their spatial distributions, into the same space, which is defined by physics rather than by the device or the behavior that produced the signal. \emph{In short, anchoring the representation to force is what produces consistency across heterogeneous skills, rather than skill-specific fitting.}

Taken together, these results support the design of Section~\ref{sec:tactile-motivation}. The force-based representation encodes the universal force attribute of touch and generalizes over a wide range of embodied manipulation behaviors, which makes it a stable carrier of tactile interaction. We expect this representation to generalize further, so that a tactile signal carries hardware-agnostic information that remains consistent across sensors and that downstream manipulation policies can consume directly.

\section{Evaluations on Tactile-aware Manipulation Tasks}
\label{sec:benchmark}

We evaluate tactile-aware manipulation on contact-rich tasks where success depends on force regulation, fine-grained contact state, deformable-object handling, delicate insertion, and tactile feedback as an efficient control signal. 
The evaluation includes NeoReal for physical robot evaluation and NeoSim for reproducible simulation evaluation. 
For both NeoReal and NeoSim, we first describe the task set and protocol, and then report the evaluation of representative policy families.

\begin{figure}[t]
    \centering
    \includegraphics[width=\linewidth]{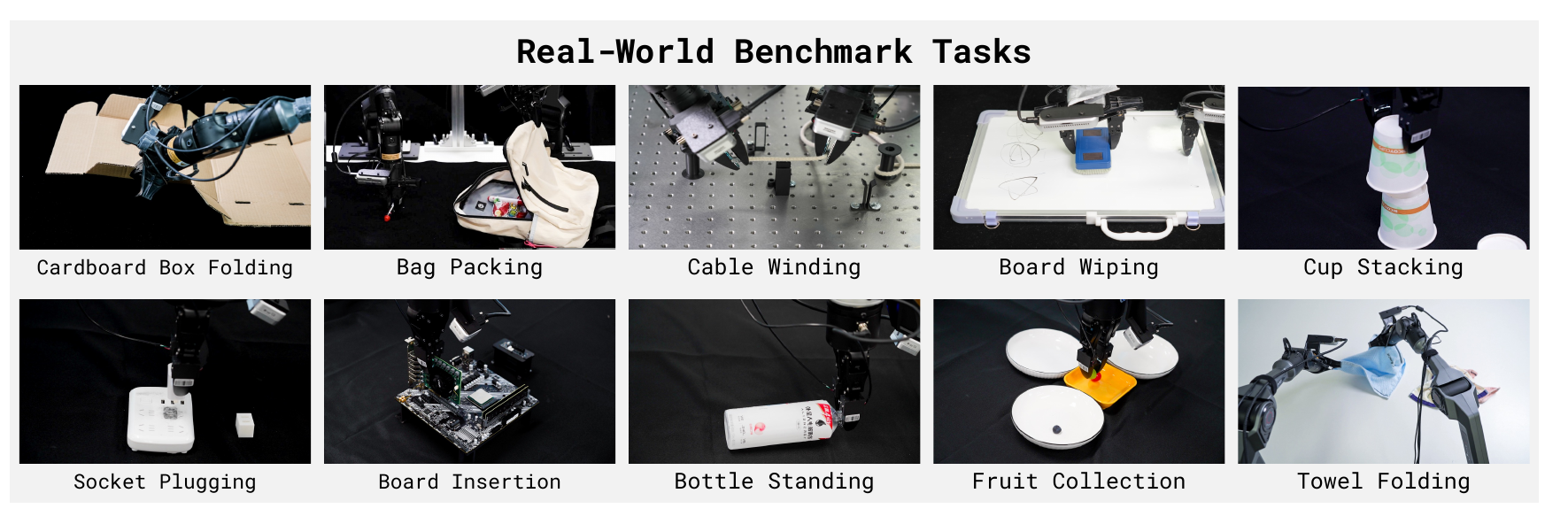}
    \caption{The NeoReal suite. NeoReal comprises 10 real-world contact-rich tasks executed on robots equipped with tactile fingers: Cardboard Box Folding, Bag Packing, Cable Winding, Board Wiping, and Cup Stacking (top row); Socket Plugging, Board Insertion, Bottle Standing, Fruit Collection, and Towel Folding (bottom row).}
    \label{fig:neoreal-tasks}
\end{figure}

\subsection{NeoReal}

NeoReal is designed to measure policy performance under fine-grained physical interaction, evaluating how policies behave under real-world tactile interaction and feedback. The suite contains 10 tasks selected from frequent and tactile-relevant skills in NeoData. These tasks cover deformable-object shaping, force-guided mating, delicate grasping, sustained surface contact, cable routing, stacking, insertion, and bimanual folding. As shown in Fig.~\ref{fig:neoreal-tasks}, the task set includes Cardboard Box Folding, Bag Packing, Cable Winding, Board Wiping, Cup Stacking, Socket Plugging, Board Insertion, Bottle Standing, Fruit Collection, and Towel Folding.

Each task specifies a standardized initial state distribution, reset protocol, and binary success criterion. The tasks are executed on robots equipped with the tactile fingers described in Section~\ref{subsec:tactile-sensor}. NeoReal therefore serves as the primary physical evaluation suite for determining whether tactile feedback improves manipulation performance in deployment. Per-task descriptions and success criteria are provided in Appendix~\ref{sec:neoreal-tasks}.

\begin{figure*}[t]
\centering
\includegraphics[width=\linewidth]{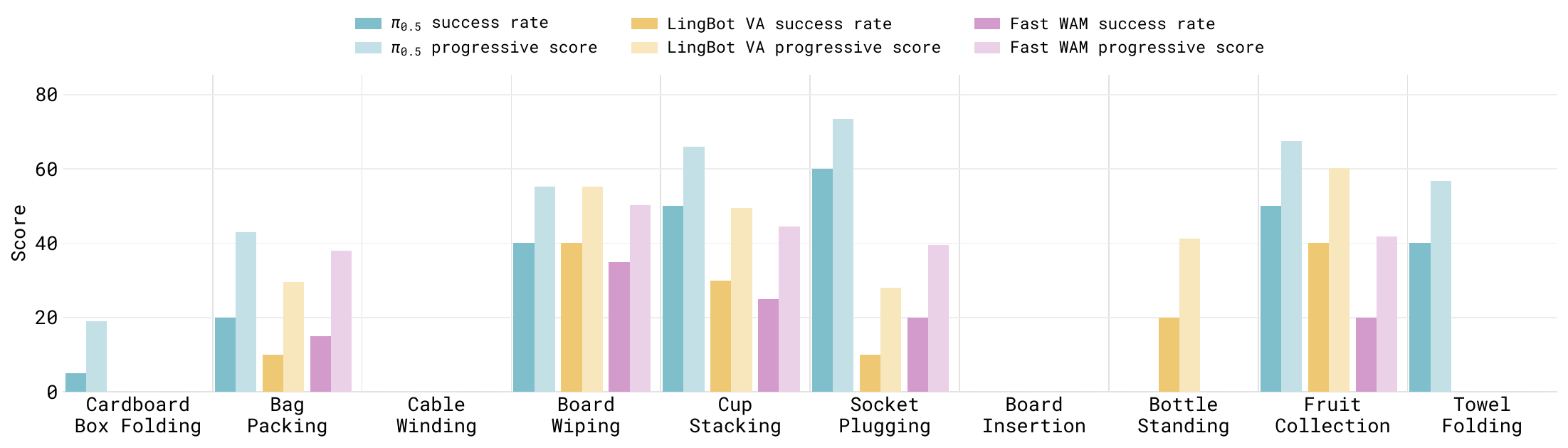}
\caption{Task-level results on the NeoReal benchmark for the three strongest policies, $\pi_{0.5}$, LingBot-VA, and Fast-WAM. For each of the ten tasks, the six bars show, per policy, the binary success rate over $20$ trials (solid) and the task progressive score (light) side by side. The progressive score credits partial completion of the per-task milestones defined in Appendix~\ref{sec:neoreal-progressive} and averages over the same $20$ trials, so it is never lower than the success rate.}
\label{fig:neoreal-results}
\end{figure*}

\subsection{Evaluation on NeoReal}
\label{sec:eval-neoreal}

\paragraph{Setup.} We evaluate representative policy families on NeoReal to validate their ability on contact-rich tactile manipulation tasks. The baselines include the imitation learning policies ACT~\citep{zhao2023learning} and Diffusion Policy~\citep{chi2023diffusion}. For VLA, we evaluate $\pi_{0.5}$~\citep{black2025pi05}, Xiaomi-Robotics-0~\citep{cai2026xiaomirobotics}, InternVLA-A1~\citep{cai2026internvla}, and StarVLA-$\alpha$~\citep{ye2026starvla}. We further include the world action models Fast-WAM~\citep{yuan2026fastwam} and GigaWorld Policy~\citep{ye2026gigaworld}, together with the VA baseline LingBot-VA~\citep{li2026causal}. For NeoReal, policies are pretrained on NeoData and then post-trained on each task, and unless otherwise stated NeoReal reports success rate over $20$ randomized rollouts per task. Full training and evaluation settings are provided in Appendices~\ref{sec:neoreal-tasks} and~\ref{sec:neosim-tasks}.

Fig.~\ref{fig:neoreal-results} reports both the binary success rate and the progressive score of the three strongest policies, $\pi_{0.5}$, LingBot-VA, and Fast-WAM, on the 10 NeoReal tasks, where the progressive score credits partial completion of the per-task milestones defined in Appendix~\ref{sec:neoreal-progressive}. Under real contact dynamics, the tasks are markedly more challenging than in simulation, and $\pi_{0.5}$ reaches the highest average success rate ($26.5\%$) by handling the sustained contact and force-guided tasks such as Board Wiping, Cup Stacking, and Socket Plugging, while LingBot-VA and Fast-WAM trail and solve only a subset of them. The progressive score consistently exceeds the success rate for every policy, and the gap widens on the contact-rich tasks, where on Socket Plugging $\pi_{0.5}$ moves from a $60\%$ success rate to a $73.5\%$ progressive score, and on Bag Packing from $20\%$ to $43.0\%$, which indicates that the policies frequently reach intermediate milestones such as grasping and alignment even when they fail to complete the task. The progressive score also preserves the overall ranking, with $\pi_{0.5}$ still leading at an average of $38.1\%$. Under both metrics, every policy retains substantial room for improvement on these tactile tasks, and the shortfall is most pronounced where touch has to act as staged feedback, such as verifying a secure grasp before lifting or a seated fit before releasing. We believe that policies driven by visual observations and proprioceptive state alone remain limited on fine and precise manipulation, since neither modality exposes the contact state to perceive, nor the signal needed to correct an action while it is still in progress.

\begin{table}[t]
\centering
\scriptsize
\setlength{\tabcolsep}{3pt}
\renewcommand{\arraystretch}{1.08}
\resizebox{\linewidth}{!}{%
\begin{tabular}{l c c c c c c c c c c c}
\toprule
Model (based on $\pi_{0.5}$) & Box & Bag & Cable & Wipe & Cups & Socket & Board & Bottle & Fruit & Towel & Avg \\
& \multicolumn{11}{c}{\footnotesize success rate / progressive score} \\
\midrule
No tactile & 5/19.0 & 20/43.0 & 0/0.0 & 40/55.2 & 50/66.0 & 60/73.5 & 0/0.0 & 0/0.0 & 50/67.5 & 40/56.8 & 26.5/38.1 \\
Tactile image concatenation & 5/17.0 & 15/38.0 & 0/0.0 & 45/58.2 & 45/61.0 & 60/71.0 & \textbf{5/19.0} & 5/20.8 & 50/69.0 & 45/59.5 & 27.5/41.4 \\
Tactile image action expert conditioning & 5/19.0 & \textbf{25/48.0} & 0/0.0 & 45/60.7 & \textbf{55/70.0} & 55/70.0 & \textbf{5/19.0} & 5/20.8 & \textbf{55/71.0} & 50/64.5 & 30.0/44.3 \\
NeoForce representation action expert conditioning & \textbf{15/34.5} & \textbf{25/48.0} & 0/0.0 & \textbf{50/63.2} & \textbf{55/70.0} & \textbf{65/76.0} & \textbf{5/19.0} & \textbf{10/29.8} & 50/67.5 & \textbf{50/66.5} & \textbf{32.5/47.5} \\
\bottomrule
\end{tabular}
}
\caption{Effect of tactile integration strategy on NeoReal, using $\pi_{0.5}$ as a fixed backbone. Each cell reports \emph{success rate / progressive score} as in Fig.~\ref{fig:neoreal-results}, and the No tactile row reproduces the $\pi_{0.5}$ result there. Column abbreviations denote Cardboard Box Folding (Box), Bag Packing (Bag), Cable Winding (Cable), Board Wiping (Wipe), Cup Stacking (Cups), Socket Plugging (Socket), Board Insertion (Board), Bottle Standing (Bottle), Fruit Collection (Fruit), and Towel Folding (Towel).}
\label{tab:tactile-integration}
\end{table}

Furthermore, we use NeoReal to analyze how tactile information should be integrated with the policy. Using $\pi_{0.5}$ as a fixed backbone, we compare four variants, namely no tactile input, raw tactile image concatenation, raw tactile image conditioning for the action expert, and NeoForce force representation conditioning for the action expert. As reported in Tab.~\ref{tab:tactile-integration}, every tactile variant outperforms the vision-only policy, confirming that contact-rich tasks benefit from information about force and contact state. Conditioning the action expert on tactile images is more effective than simple concatenation, because the contact signal is supplied directly to the module that generates actions. The NeoForce representation improves further over raw tactile images, since it expresses contact as a compact force field rather than as device appearance. The average progressive score follows the same ordering, rising from $38.1\%$ for the vision-only baseline to $47.5\%$ for NeoForce conditioning, and although individual tasks fluctuate, with the raw tactile variants for instance dropping on Bag Packing and Cup Stacking before the force representation recovers them, the aggregate trend confirms that the gains also appear as more complete milestone progress and not only as additional binary successes. \emph{In short, a policy benefits from the physical contact representation rather than from the device-specific appearance, and a sensor-agnostic force field provides a universal interface that delivers such a representation.}

\begin{figure}[t]
    \centering
    \includegraphics[width=\linewidth]{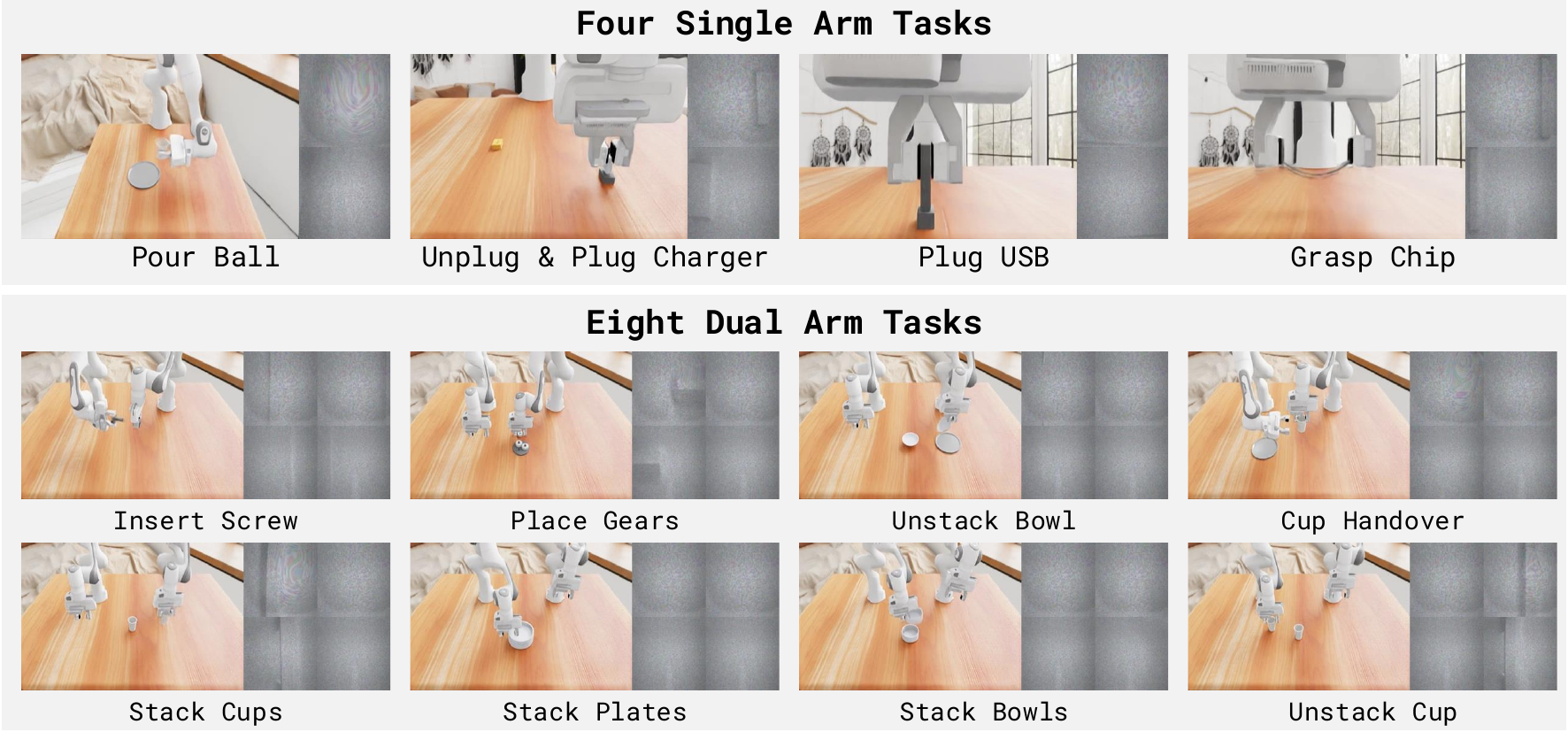}
    \caption{Demonstrations from NeoSim. NeoSim contains $12$ tasks, including $4$ single-arm tasks and $8$ dual-arm tasks. For each task, the left panel shows the simulated RGB scene and the right panel visualizes the particle gel images rendered from the tactile sensors.}
    \label{fig:benchmark-neosim-demonstration}
\end{figure}

\subsection{NeoSim}

NeoSim is constructed in the simulation environment and provides a reproducible arena for tactile-aware evaluation at scale. Following UniVTAC~\citep{chen2026univtac}, NeoSim performs high-precision tactile data collection in simulation and renders per-contact force fields in the same $x$, $y$, and $z$ format as the unified representation in Section~\ref{sec:tactile-representation}. This design allows policies that consume force-based tactile inputs to be evaluated in simulation with the same representation used for real tactile data.

The suite contains $12$ contact-rich tasks, including $4$ single-arm tasks and $8$ dual-arm tasks. The single-arm tasks are Pour Ball, Unplug and Plug Charger, Plug USB, and Grasp Chip, which stress force regulation during pouring, insertion, and delicate grasping. The dual-arm tasks are Insert Screw, Place Gears, Unstack Bowl, Cup Handover, Stack Cups, Stack Plates, Stack Bowls, and Unstack Cup, which further require bimanual coordination, stable contact maintenance, handover, and nesting. These tasks span rigid objects, thin-walled containers, fragile parts, deformable contact, and tight-clearance assembly.

NeoSim is intended to complement NeoReal by providing controlled, reproducible, and large-scale evaluation that is difficult to obtain on physical hardware. It enables controlled comparisons across policy families, tactile input choices, and representation choices under large-scale contact dynamics. A detailed per-task description of NeoSim, together with rendered examples of RGB observations and tactile force fields, is provided in Appendix~\ref{sec:neosim-tasks}.

\begin{table}[t]
\centering
\scriptsize
\setlength{\tabcolsep}{3pt}
\renewcommand{\arraystretch}{1.08}
\resizebox{\linewidth}{!}{%
\begin{tabular}{l c c c c c c c c c c c c c}
\toprule
\multirow{2}{*}{Policy}
& \multicolumn{4}{c}{Single arm}
& \multicolumn{8}{c}{Dual arm}
& \multirow{2}{*}{Mean} \\
\cmidrule(lr){2-5}\cmidrule(lr){6-13}
& USB & Chip & Charger & Pour
& Bowl U & Cup S & Screw & Gears & Cup H & Plate S & Bowl S & Cup U & \\
\midrule
$\pi_{0.5}$~\citep{black2025pi05} & \textbf{74} & 86 & \textbf{23} & \textbf{92} & 3 & \textbf{22} & \textbf{26} & \textbf{18} & 13 & \textbf{96} & \textbf{93} & 3 & \textbf{45.8} \\
Xiaomi-Robotics-0~\citep{cai2026xiaomirobotics} & 62 & 49 & 5 & 52 & 0 & 0 & 0 & 0 & 0 & 0 & 42 & \textbf{71} & 23.4 \\
InternVLA-A1~\citep{cai2026internvla} & 36 & 12 & 0 & 47 & 0 & 0 & 0 & 0 & 0 & 0 & 8 & 0 & 8.6 \\
StarVLA-$\alpha$~\citep{ye2026starvla} & 72 & \textbf{93} & 0 & 32 & 17 & 0 & 8 & 0 & \textbf{25} & 0 & 0 & 31 & 23.2 \\
Fast-WAM~\citep{yuan2026fastwam} & 0 & 0 & 0 & 0 & 0 & 0 & 0 & 0 & 0 & 0 & 0 & 0 & 0.0 \\
GigaWorld Policy~\citep{ye2026gigaworld} & 34 & 71 & 3 & 0 & 0 & 12 & 0 & 0 & 0 & 0 & 0 & 9 & 10.8 \\
LingBot-VA~\citep{li2026causal} & 31 & 77 & 0 & 37 & \textbf{18} & 20 & 18 & 0 & 0 & 93 & 67 & 24 & 32.1 \\
\bottomrule
\end{tabular}
}
\caption{Task level success rates over 100 randomized rollouts on NeoSim. Column abbreviations denote Insert USB (USB), Grasp Chip (Chip), Unplug and Plug Charger (Charger), Pour Ball (Pour), Bowl Unstack (Bowl U), Cup Stack (Cup S), Insert Screw (Screw), Place Gears (Gears), Cup Handover (Cup H), Plate Stack (Plate S), Bowl Stack (Bowl S), and Cup Unstack (Cup U). Mean averages over the twelve task success rates in each row.}
\label{tab:main-results}
\end{table}

\subsection{Evaluation on NeoSim}
\label{sec:neosim-results}

We evaluate on NeoSim how current policy families handle contact-rich manipulation in a reproducible simulated setting. Using the same policy families and NeoData pretraining as above, NeoSim reports success rate over $100$ randomized rollouts per task. Tab.~\ref{tab:main-results} reports the NeoSim results. $\pi_{0.5}$ leads with a mean success rate of $45.8\%$, followed by LingBot-VA at $32.1\%$, while Xiaomi-Robotics-0 and StarVLA-$\alpha$ form a middle group near $23\%$ and the remaining baselines stay below $11\%$. The dual-arm tasks are consistently more difficult than the single-arm tasks. Averaged over the four single-arm tasks, every policy scores higher than it does over the eight dual-arm tasks, and the drop is severe for most of them, from $49.3\%$ to $10.1\%$ for StarVLA-$\alpha$, from $42.0\%$ to $14.1\%$ for Xiaomi-Robotics-0, and from $23.8\%$ to $1.0\%$ for InternVLA-A1. Only $\pi_{0.5}$ and LingBot-VA retain substantial bimanual competence, at $34.3\%$ and $30.0\%$, respectively, and they do so almost entirely on Plate Stack and Bowl Stack, where the two arms can act in a loosely coupled manner. The tasks that demand sustained mutual contact remain nearly unsolved for every policy, with the best score reaching only $18\%$ on Place Gears and $25\%$ on Cup Handover, and on each of these two tasks at most two of the seven policies succeed at all. Fast-WAM fails on all twelve tasks despite being one of the three strongest policies on NeoReal, which indicates that its competence does not transfer to the simulated contact dynamics. These results expose contact-rich bimanual manipulation as the main bottleneck for current policy families that rely primarily on visual observations, which reveal little about whether a grasp is secure and whether contact is stably maintained. \emph{In short, tactile signals make grasp security and contact stability directly observable to the policy, which is exactly the information that contact-rich bimanual manipulation currently fails on.} This in turn strengthens the necessity of equipping robots with tactile hardware and of policies that consume a tactile representation.

\section{Conclusion}
\label{sec:conclusion}

We presented $\mathcal{N}_0$-Foundation, an integrated resource for tactile-enabled embodied manipulation that spans hardware, data, representation, and evaluation. The report contributes NeoData, a large-scale multimodal dataset with more than $30{,}000$ hours of synchronized visual and tactile demonstrations across six embodiments and $450+$ tasks, collected through a mixture of real-robot teleoperation and UMI demonstrations. It also contributes NeoForce, which convert sensor signals into tactile representation for downstream embodied models. Finally, we propose a standardized evaluation that combines the real-world NeoReal suite and the simulated NeoSim suite for standardized evaluation of contact-rich manipulation. The experiments indicate that tactile feedback improves contact-rich manipulation, that latent supervision improves temporal tactile representation learning, and that the NeoForce force representation provides a compact physical alternative to raw tactile images.

Several directions remain open. The unified representation currently targets parallel-jaw visuo-tactile fingers, and extending it to dexterous hands and non-camera-based transducers would broaden its applicability. Scaling policy learning to the full multimodal corpus and tightening the connection between NeoSim and real deployment are natural next steps. NeoData, together with its representation and benchmark, provides a foundation for more capable and transferable tactile-aware manipulation policies.

\section*{Contributors}

\textbf{Data Infrastructure.} Xiufeng Song, Rui Li, Wenjie Zhou, Yutao Fan, Heng Zhou, Tianyu Yang, Longjie Su, Li Kang, Zhemeng Zhang, Kun Huang, Haotian Wang, Shitao Lin, Shunlin Lu, Yiran Qin.\\
\textbf{Data Preprocess and Annotation.} Rui Li, Xiufeng Song, Tianyu Yang, Silong Dai, Longjie Su.\\   
\textbf{Evaluation.} Bruno N.Y. Chen, Jiongwei Lu, Yifan Wang, Shengqi Xu, Yutao Fan, Zipei Ma, Xiaofei Wei, Zhemeng Zhang, Heng Zhou, Silong Dai, Li Kang, Longjie Su, Boyu Mi, Yanjun Li, Chenxi Wu, Nan Min, Guojin Zhong, Haoyu Zhao, Xiufeng Song, Rui Li, Xin Wang, Yiran Qin.\\
\textbf{Hardware.} Hu Luo, Liuchuang Tang, Yu Han, Hua Zhu, Yutao Ling, Qiuhui Huang, Fanjie Wang, Lianjie Jiang, Hao Chen, Yicheng Yang, Yinxiao Lu.\\
\textbf{Academic Supervision.} Ziyi Ye, Guoxiang Dong, Xiaosong Jia, Yanwei Fu, Wenming Chen. \\
\textbf{Project Lead.} Shunlin Lu, Yiran Qin, Shihao Zhao, Daoguo Dong, Zuxuan Wu, Yu-Gang Jiang.

\bibliography{main}

\newpage
\appendix

\section{NeoData Information}
\label{sec:supp-data}

\paragraph{Scale and Episodes.} NeoData comprises approximately $1.4$M episodes totaling more than $30{,}000$ hours of interaction, $3.3$B timesteps, $8$B RGB frames, and $10$B tactile frames, collected across six embodiments by more than $90$ human operators. The mean episode duration is about $80$ seconds, and the distribution of episode lengths is approximately unimodal, as reported in Fig.~\ref{fig:neodata-statistics}. The object-frequency distribution is long-tailed, with cups and boxes followed by test tubes and building blocks among the most frequent categories. Task frequencies are summarized separately in Fig.~\ref{fig:neodata-task-classification-bar}, where a small set of high-volume skills, led by cloth folding, coexists with a broad tail of rarer tasks.

\paragraph{Data Definition.} Every episode is a synchronized, temporally aligned stream of observations and actions. UMI observations contain one fisheye wrist RGB image and two visuo-tactile images from the left and right fingers, and all three streams are stored at a native resolution of $640 \times 360$. Robot embodiment observations additionally include external RGB views when available. Following Section~\ref{subsec:data-format}, actions are stored as relative end-effector commands and gripper width for UMI data, and as tool center point commands together with absolute joint commands for robot embodiment data. Alongside the raw tactile images, each episode also carries the derived NeoForce force field at $640 \times 360$ resolution.

\paragraph{Task Types.} The corpus spans more than $450$ tasks, ranging from everyday household activities such as folding clothes, folding pants, wiping, and pick-and-place, to delicate contact-rich industrial operations such as small part assembly, screw tightening, USB and charger plugging, and accessory classification. As summarized in Fig.~\ref{fig:neodata-task-classification-bar}, the most common skill categories are folding, pick-and-place, and wiping or cleaning, followed by a long tail of contact-rich primitives including stacking, assembling, pouring, inserting, plugging, and pressing. The tasks are distributed across six scene types and a diverse object set grouped into containers, food and lab items, tools, deformables, kitchenware, and electronics. Their scene and leading object distributions are shown in Fig.~\ref{fig:neodata-statistics}.

\section{$\mathcal{N}_0$-TacUMI Details}
\label{sec:supp-tacumi}

\paragraph{Collection Device.} The robot-free portion of NeoData is gathered with the $\mathcal{N}_0$-TacUMI device introduced in Section~\ref{subsec:tactile-umi}, a handheld parallel gripper carrying two tactile sensors, an infrared tracker for six-degree-of-freedom pose, and a magnetic encoder for finger separation. All streams are recorded synchronously at $30$~fps against a common clock, and the same sensors and schema are reused on the teleoperated robots so that the two regimes are directly interoperable.

\paragraph{Action.} At each timestep the tracker reports the gripper pose $P_t \in \mathrm{SE}(3)$ and the magnetic encoder reports the gripper width $g_t$. Rather than storing absolute poses as targets, we record the relative transform to a frame $k$ steps ahead, $\Delta P_t = P_t^{-1} P_{t+k}$, and define the action as $a_t = (\Delta P_t, g_{t+k})$, pairing the relative end-effector motion with the commanded gripper width. This relative pose convention is invariant to the world frame and supports transfer across embodiments.

\paragraph{Wrist Images.} The wrist view is captured by the fisheye RGB camera mounted on the handheld gripper, at $640 \times 360$ and $30$~fps, temporally aligned with the pose and tactile streams. It provides the egocentric visual context used as a policy input and as the visual stream that NeoForce fuses with the tactile force field.

\paragraph{Tactile Images.} Each finger embeds a camera-based visuo-tactile sensor (Section~\ref{sec:tactile-representation}) that images the deformation of its sensing surface. The left and right tactile images $T_t^{l}, T_t^{r}$ are recorded at $640 \times 360$ and $30$~fps in lockstep with the wrist image and the action. During preprocessing, each tactile image is mapped through the learned inverse model $g_\theta$ into the $640 \times 360$ force field, yielding the unified tactile representation stored alongside the raw frames.

\section{Annotation Pipeline Details}
\label{sec:supp-annotation}

We build a semi-automatic hierarchical annotation pipeline that labels every robot manipulation video with a four-level, temporally nested structure of task (L3), subtask (L2), action (L1), and atomic segment (L0). The pipeline runs in two stages, a task template generation stage and a per-episode annotation stage.

\paragraph{Task Template Construction}
For each task category, we first use the video summarization capability of a VLM, Gemini-3.5-Flash, to generate a time-independent task template from a few representative episodes. The template specifies the L3 overall objective shared by the task, an ordered set of L2 semantic phases, and the fixed L1 substeps under each phase. A human expert then reviews and confirms this template. This review is the only manual quality control step in the entire pipeline, and it removes the semantic wording drift that would otherwise arise across different episodes of the same task.

\paragraph{Episode Segmentation}
After the template is confirmed, we annotate each episode in a boundary-first, label-second order. The lowest level L0 atomic intervals are first delimited by two parallel branches that operate on two complementary modalities. The action signal branch locates a set of physically grounded atomic action boundaries from the kinematic signals of the end effector and gripper together with our self-collected tactile signals. These include the grasp boundary, whose criterion is that the gripper transitions from open to closed while the tactile signal rises sharply, indicating that an object is now held, with the remaining boundaries using analogous criteria, as well as the release boundary, the motion start and stop boundaries, the axis switch boundary, the vertical reversal boundary, and the contact and separation boundaries. Each boundary is aligned precisely to the moment at which the mechanical state changes. The visual boundary branch applies unsupervised Generic Event Boundary Detection (GEBD)~\cite{gebd}, a method that perceives event boundaries from the visual stream. We subsample the video at $4$~fps, extract ResNet-50~\cite{resnet} features pretrained on ImageNet~\cite{imagenet}, and at each timestep compare the predictability, measured by cosine distance, between the average features of the $5$ preceding and the $5$ following frames, taking local maxima at points of abrupt semantic change as boundaries. This branch captures perceptual action transitions that the control signals cannot encode, such as the transition from folding to wiping that carries no clear mechanical discontinuity. The two boundary sets are then fused and deduplicated, and the total number of segments is capped at $24$ while fine granularity is preserved, yielding a temporally fixed and more accurately partitioned set of L0 atomic intervals.

\paragraph{Hierarchical Labeling}
Finally, we let the same VLM, Gemini-3.5-Flash, label only these pre-segmented intervals. It assigns each L0 segment to the corresponding L1 and L2 level in the template and generates the associated description, after which deterministic code assembles the temporal extents of the L1, L2, and L3 levels from the bottom up. The temporal endpoints at all levels come from signal computation rather than VLM inference, so this design guarantees the accuracy of the temporal intervals at every level and fundamentally avoids the temporal drift of VLMs on long videos.

\section{Force Field Conversion and NeoForce Details}
\label{sec:supp-force-pipeline}

\subsection{Real Force Calibration Data}
We collect real tactile data and calibrate the applied force to build the training data for the tactile model. The calibration corpus is produced by a controlled annotation and curation pipeline rather than by uncalibrated in the wild demonstrations. As the acquisition source, a six-degree-of-freedom robot arm presses an indenter against the tactile surface while a six axis force and torque sensor records the reference load under a quasi-static protocol, in which the indenter is held stationary once it reaches a target depth so that the tactile image and the measured force are captured in a stable state. Under automated control, the indenter traverses the entire sensing plane at penetration depths from $1$ to $6\,\mathrm{mm}$ in $1\,\mathrm{mm}$ increments, and shear supervision is enriched by tilting the indenter through rotation angles from $1^{\circ}$ to $20^{\circ}$ to cover diverse horizontal loads. Each curated sample takes the form of an optical flow image paired with a pixel aligned multi dimensional force field, where the two in-plane components are combined into a resultant shear stress and the normal component records pressure. The pipeline yields approximately $4{,}300$ optical flow and multi dimensional force pairs, which are randomly shuffled and split into training, validation, and test partitions with an $80\%$, $10\%$, and $10\%$ ratio, with the test partition held out exclusively for final evaluation.

\begin{figure}[t]
    \centering
    \includegraphics[width=0.92\linewidth]{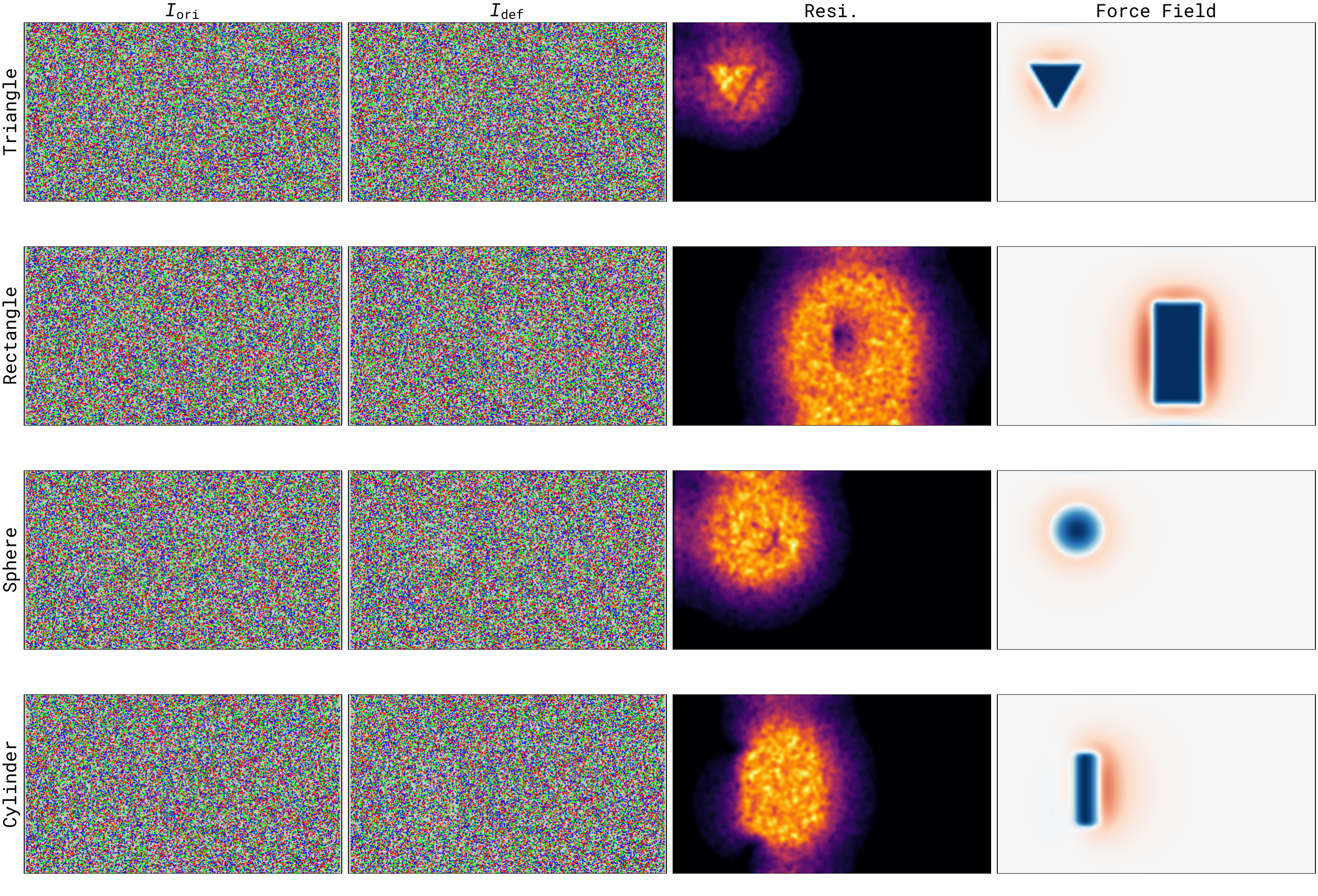}
    \caption{Simulated visuo-tactile data generation for four indenter shapes: triangle, rectangle, sphere, and cylinder. From left to right, the columns show the undeformed reference image ($I_{\mathrm{ori}}$), the deformed image after loading ($I_{\mathrm{def}}$), the residual between them (Resi.) that visualizes how the gel particles move at the contact region, and the resulting three-axis force fields. All four shapes are converted into pixel-aligned force fields, showing that the simulation pipeline produces consistent force supervision across diverse contact geometries.}
    \label{fig:supp-sim-tactile}
\end{figure}

\subsection{Tactile Simulation Augmentation}
To enlarge the diversity of contact geometries beyond what physical indentation can cover, we augment the real calibration corpus with simulated visuo-tactile data. The goal of the simulation is to synthesize supervised pairs that map visual deformation to a contact force field. Taking the relative contact position between the object and the sensing layer as the boundary condition, and combining the material and geometric parameters of the sensing layer, we drive an Abaqus finite element solver to obtain the distributed force and displacement fields. The displacement field then deforms the reference particle texture of the sensing surface through a renderer, producing an image pair before and after loading. Formally, each simulated sample is a triplet $(I_{\mathrm{ori}}, I_{\mathrm{def}}, F)$, where $I_{\mathrm{ori}} \in \mathbb{R}^{1440 \times 2560 \times 3}$ is the undeformed reference image, $I_{\mathrm{def}} \in \mathbb{R}^{1440 \times 2560 \times 3}$ is the deformed image rendered under the solved displacement field, and $F \in \mathbb{R}^{480 \times 640 \times 3}$ is the pixel aligned force field whose three channels store the $x$ shear, the $y$ shear, and the normal pressure. Since $F$ shares the image pixel coordinates, the spatial mapping is fixed by the physical size of the sensing layer, with a physical length of $28.0\,\mathrm{mm}$ spanning $2560$ pixels and a physical width of $16.0\,\mathrm{mm}$ spanning $1440$ pixels. As shown in Fig.~\ref{fig:supp-sim-tactile}, the residual between $I_{\mathrm{ori}}$ and $I_{\mathrm{def}}$ visualizes the particle motion induced by contact, and the pipeline converts this deformation into a force field for each of the four indenter shapes, so that the augmented data follows the same force field form as the real calibration corpus.

\begin{figure}[t]
    \centering
    \begin{subfigure}{0.245\linewidth}
        \includegraphics[width=\linewidth]{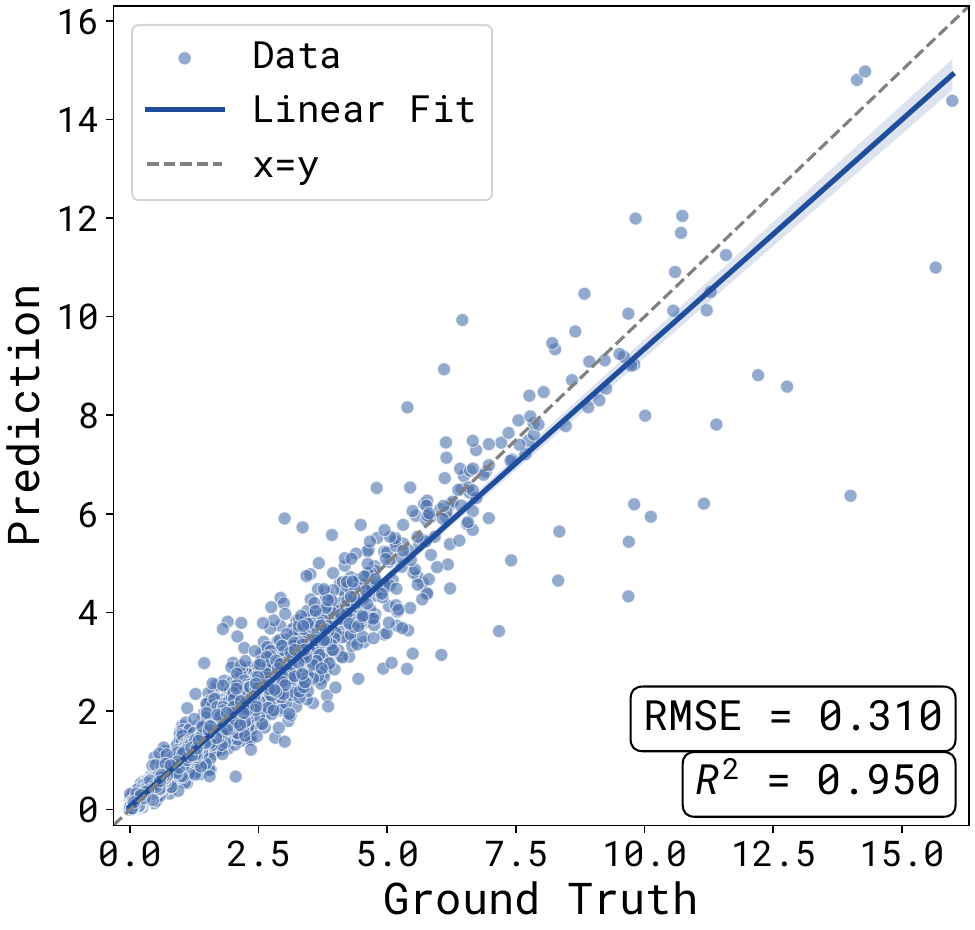}
        \caption{Pressure (real)}
    \end{subfigure}
    \hfill
    \begin{subfigure}{0.245\linewidth}
        \includegraphics[width=\linewidth]{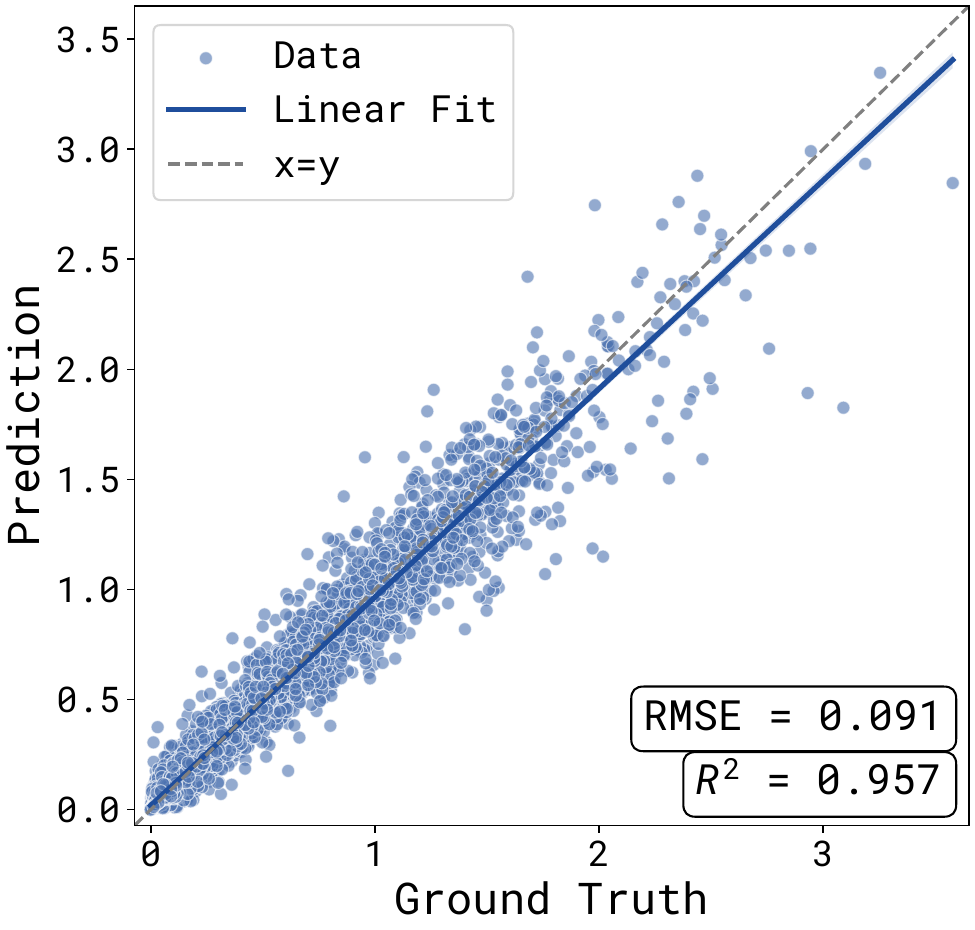}
        \caption{Shear (real)}
    \end{subfigure}
    \hfill
    \begin{subfigure}{0.245\linewidth}
        \includegraphics[width=\linewidth]{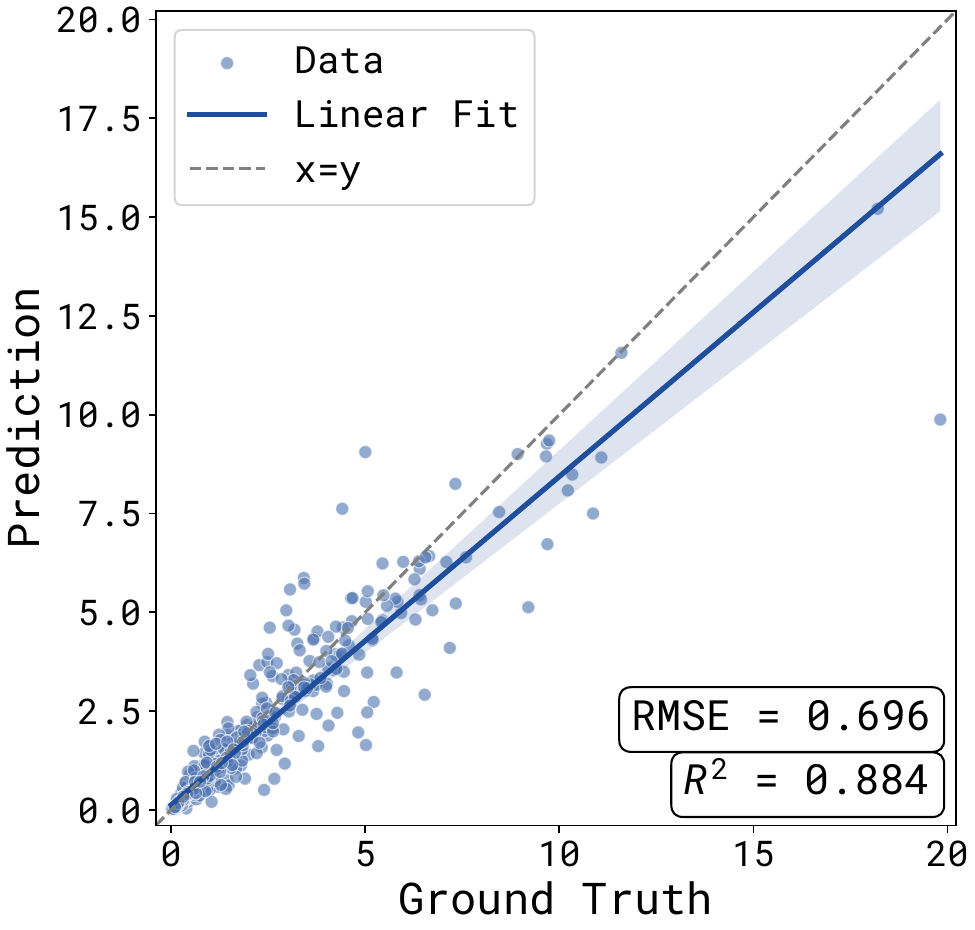}
        \caption{Pressure (sim)}
    \end{subfigure}
    \hfill
    \begin{subfigure}{0.245\linewidth}
        \includegraphics[width=\linewidth]{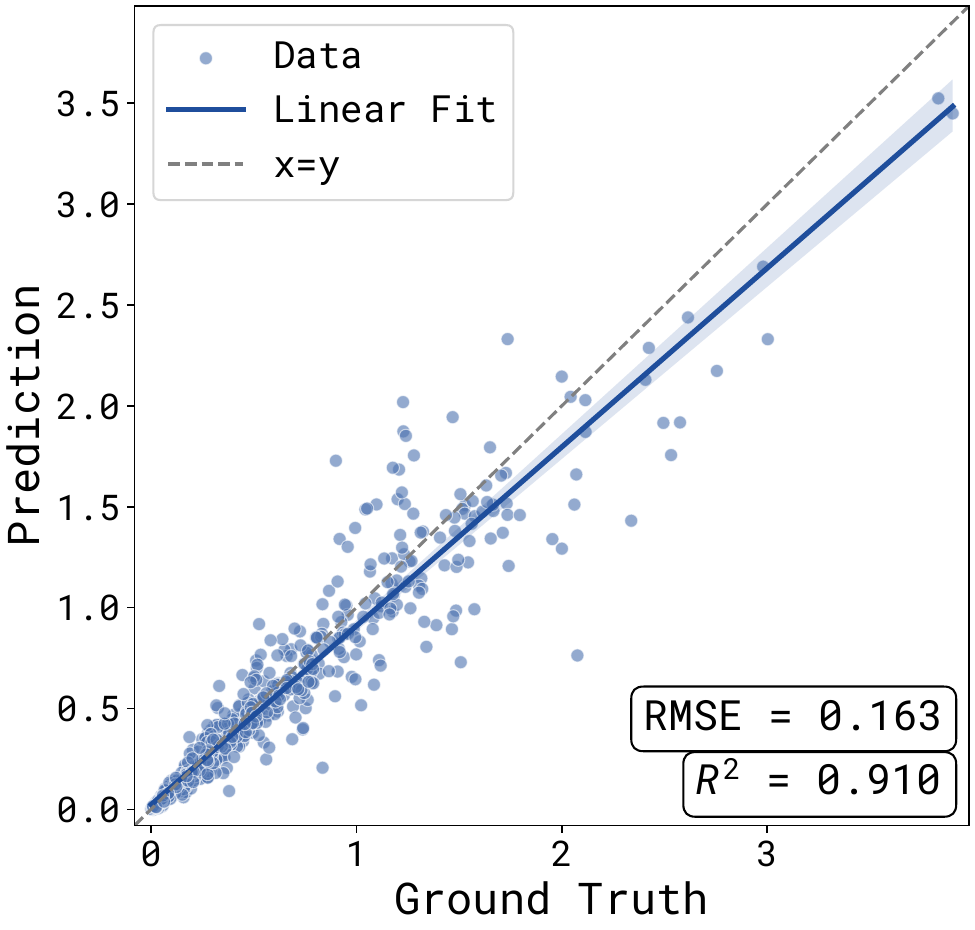}
        \caption{Shear (sim)}
    \end{subfigure}
    \caption{Evaluation of the force field conversion model. Each panel regresses the predicted force against the ground truth with a linear fit and the $x=y$ reference line, reporting RMSE and $R^2$, for pressure and shear in the real setting and on the simulated data.}
    \label{fig:force-regression}
\end{figure}

\subsection{Force Field Conversion Model}
The tactile image to force field converter adopts GoogLeNet~\citep{googlenet} as the backbone, whose inception modules aggregate multiscale features at a controlled computational cost, and modifies the output layer to regress the three-axis per-pixel force field that aligns with the shear and pressure targets. The converter is trained on a mixture of the real calibration corpus and the simulation augmented data described above, so that it benefits from both physically measured contact and the broader shape diversity of simulation. Following the reference training protocol, the model is implemented in PyTorch and optimized with AdamW using a batch size of $25$, an initial learning rate of $0.001$ decayed by a factor of $10$ every $10$ epochs, for a total of $50$ epochs. The objective combines an $L_1$ per-pixel force regression loss with an $L_2$ contact segmentation loss, each normalized by the number of valid contact pixels so that the loss reflects the average per-pixel error and is robust to the varying size of the contact region. As reported in Fig.~\ref{fig:force-regression}, the converter regresses force accurately in the real setting, reaching $R^2 = 0.950$ with RMSE $= 0.310$ for pressure and $R^2 = 0.957$ with RMSE $= 0.091$ for shear, and it remains reliable on the simulated data, reaching $R^2 = 0.884$ with RMSE $= 0.696$ for pressure and $R^2 = 0.910$ with RMSE $= 0.163$ for shear. The consistently high linearity and low error across both domains and both force components confirm that the converter performs effective force regression and can serve as the annotation model that labels the large-scale tactile corpus with physically meaningful force fields.

\subsection{NeoForce Experiment Details}
We train NeoForce on $20{,}000$ real-robot and $\mathcal{N}_0$-TacUMI demonstrations spanning $30$ tasks and evaluate it on $2{,}500$ held-out episodes, which endows the model with broad contact-rich manipulation priors. Each training sample is a chunk of synchronized RGB observations and tactile force fields, and the chunk length is set to $4$ in our experiments. The visual and tactile streams are patchified independently and then jointly processed by a shared transformer backbone, for which we adopt a ViT-B initialized from DINOv2 pretrained weights, on top of which sit two prediction heads. The reconstruction head decodes force fields and contact masks, while the latent prediction head is supervised by an exponential moving average teacher whose output serves as the ground truth in latent space. Given masked inputs, the student predicts these teacher latents for visual tokens, tactile tokens, cross-modal alignment, and masked patch tokens. The complete training objective follows Section~\ref{sec:tactile-representation}. We train NeoForce on $8\times$ NVIDIA A100 GPUs for $100{,}000$ steps with a learning rate of $2\times10^{-5}$.

\section{NeoReal Task Suite}
\label{sec:neoreal-tasks}

For each task, we provide a representative benchmark image and describe the manipulation skills and tactile capabilities it evaluates. NeoReal contains 10 real-world contact-rich manipulation tasks executed on physical robot setups equipped with the tactile fingers described in Section~\ref{subsec:tactile-sensor}. Each task uses a standardized initial state distribution and reset protocol, and success is judged by a binary task-specific criterion such as complete insertion, stable stacking, correct routing, object placement, or final shape quality. For each task we show the real benchmark image extracted from the NeoReal task suite slide and describe the manipulation skill it exercises.

\subsection{Data Demonstrations}
\label{sec:neoreal-demos}

\par\vspace{0.6em}\noindent
\begin{minipage}[c]{0.44\linewidth}
  \includegraphics[width=\linewidth]{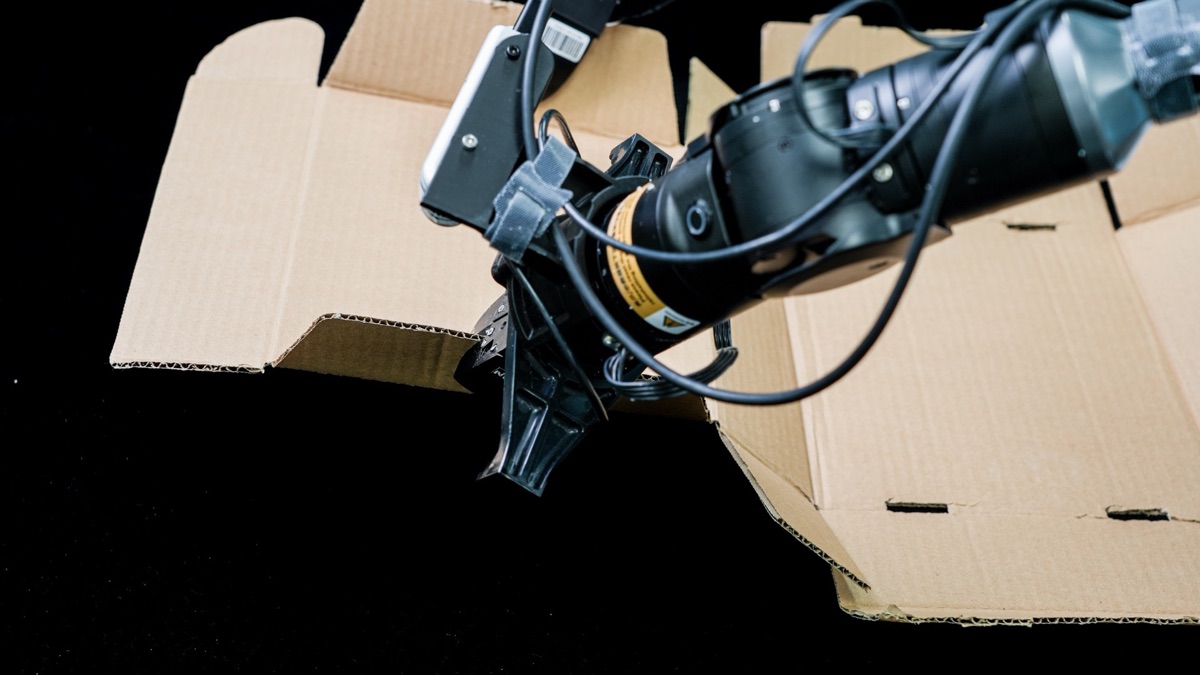}
\end{minipage}\hfill
\begin{minipage}[c]{0.52\linewidth}
  \textbf{Cardboard Box Folding.} The robot manipulates a cardboard box blank and folds it into the target box shape. The task stresses contact-rich bending, crease following, and force controlled pressing so that the cardboard seats correctly without tearing or collapsing.
\end{minipage}\par\vspace{0.6em}

\par\vspace{0.6em}\noindent
\begin{minipage}[c]{0.44\linewidth}
  \includegraphics[width=\linewidth]{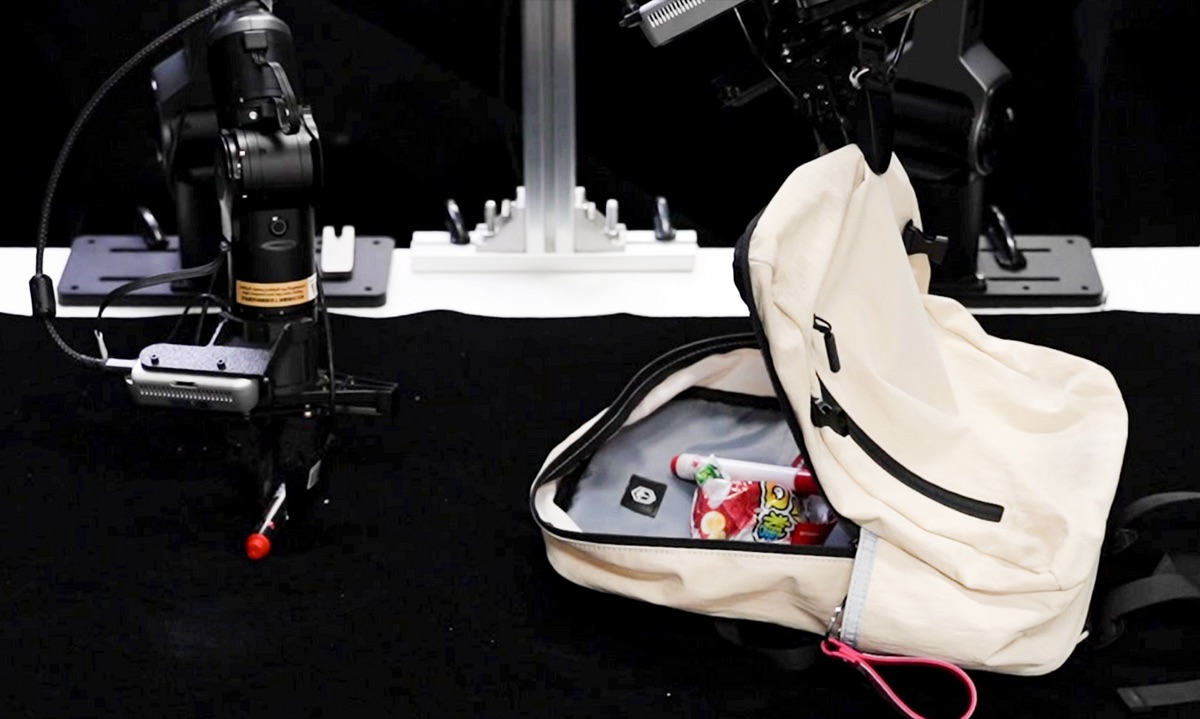}
\end{minipage}\hfill
\begin{minipage}[c]{0.52\linewidth}
  \textbf{Bag Packing.} The robot picks up the scattered items on the table, packs them into the bag, and pulls the bag zipper closed. The task combines cluttered-object grasping, placement inside a compliant container, and contact-rich zipper closing that requires sensing the pulling resistance along the zipper track.
\end{minipage}\par\vspace{0.6em}

\par\vspace{0.6em}\noindent
\begin{minipage}[c]{0.44\linewidth}
  \includegraphics[width=\linewidth]{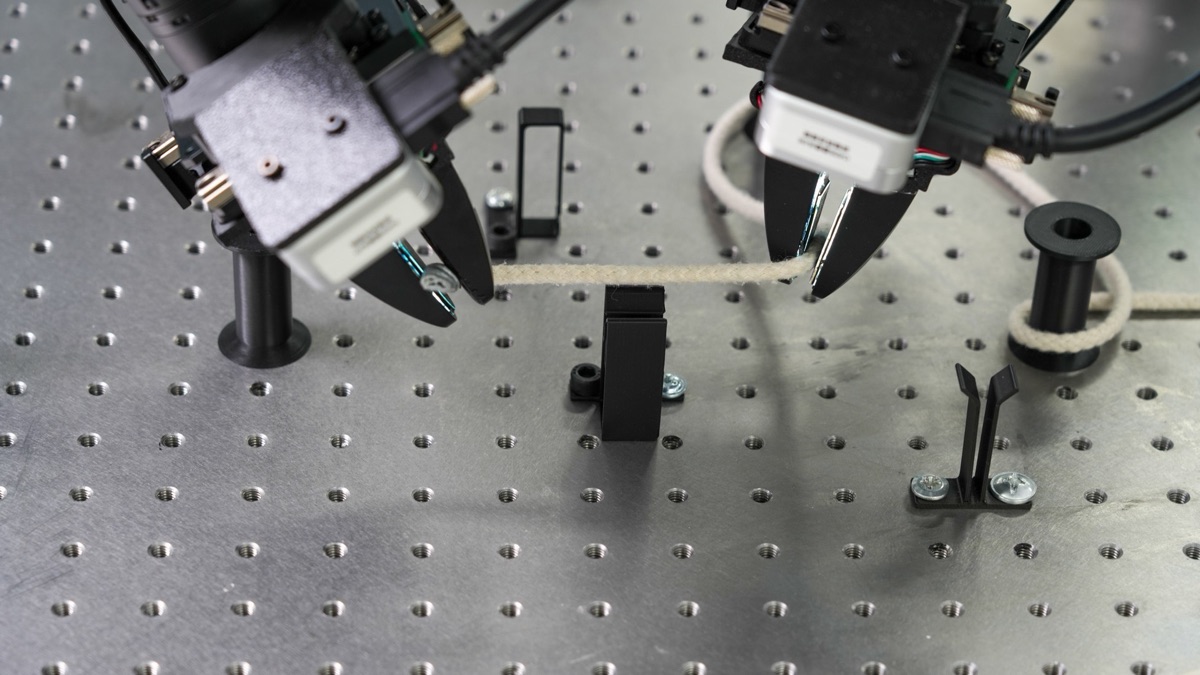}
\end{minipage}\hfill
\begin{minipage}[c]{0.52\linewidth}
  \textbf{Cable Winding.} The robot routes and winds a cable around the designated fixtures. The task requires maintaining cable tension, following the target path, and avoiding missed posts, tangles, or crossing errors.
\end{minipage}\par\vspace{0.6em}

\par\vspace{0.6em}\noindent
\begin{minipage}[c]{0.44\linewidth}
  \includegraphics[width=\linewidth]{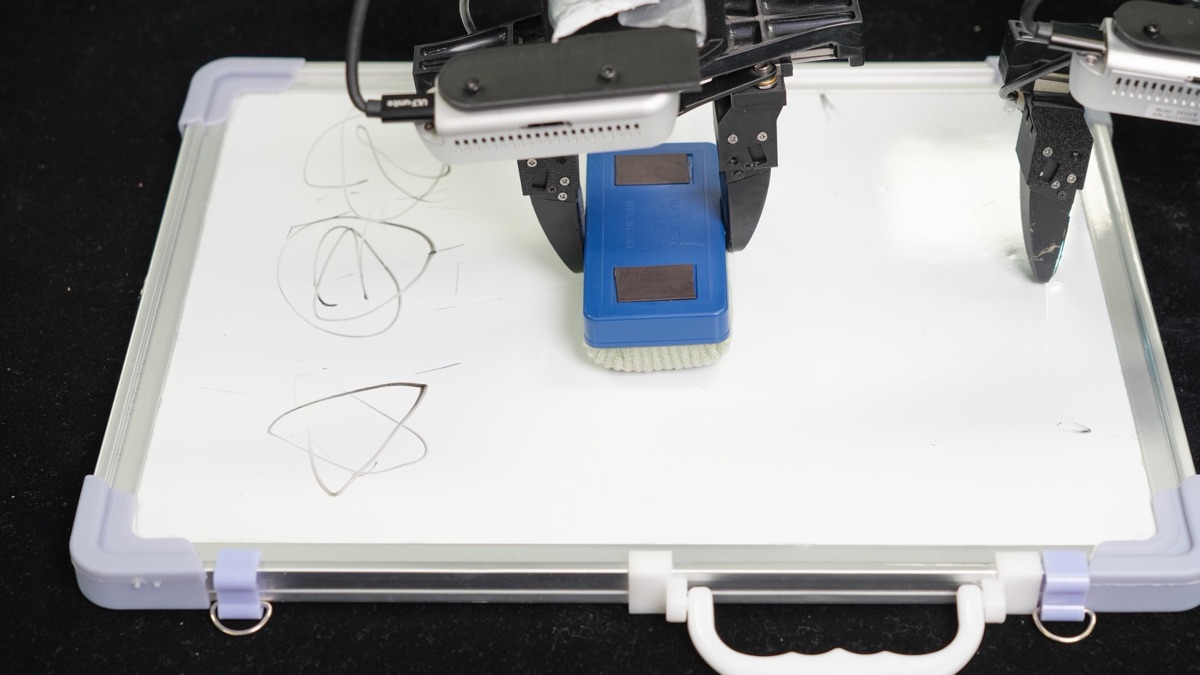}
\end{minipage}\hfill
\begin{minipage}[c]{0.52\linewidth}
  \textbf{Board Wiping.} The robot presses a wiping tool against a whiteboard and cleans the marked region. The task tests sustained surface contact, stable normal force, and coverage of the target area without losing contact or pushing the board out of place.
\end{minipage}\par\vspace{0.6em}

\par\vspace{0.6em}\noindent
\begin{minipage}[c]{0.44\linewidth}
  \includegraphics[width=\linewidth]{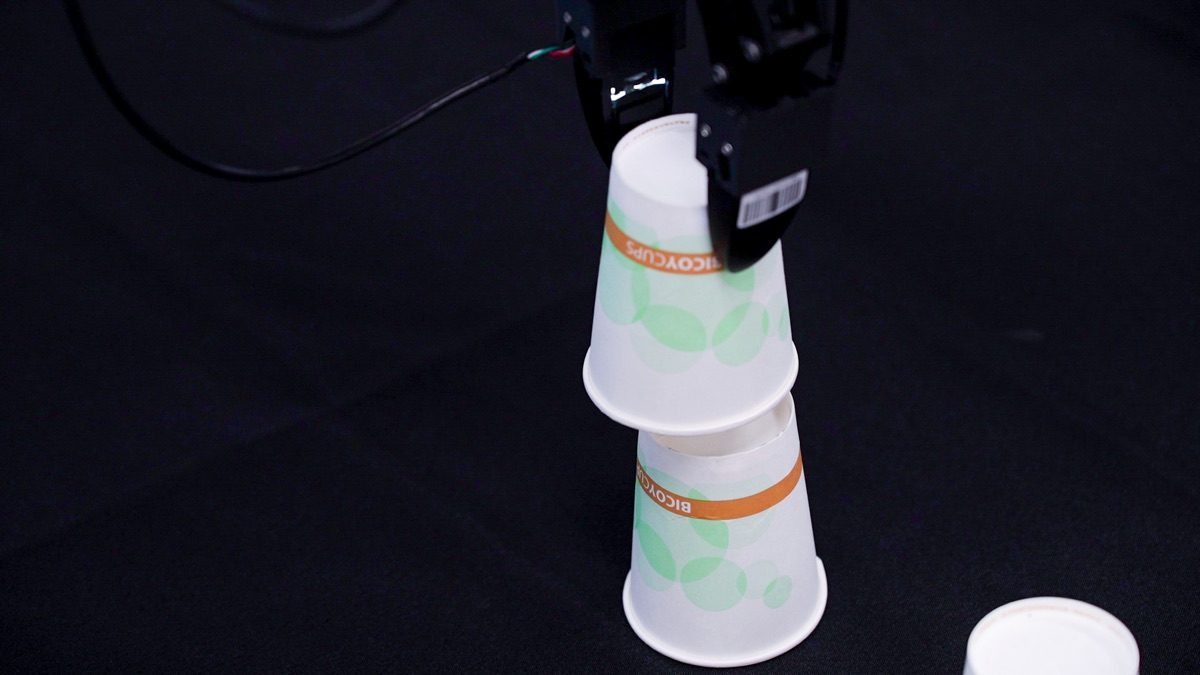}
\end{minipage}\hfill
\begin{minipage}[c]{0.52\linewidth}
  \textbf{Cup Stacking.} The robot stacks lightweight cups into a stable nested arrangement. The cups are thin-walled and easily deformed, so the policy must align them while limiting contact force during insertion and seating.
\end{minipage}\par\vspace{0.6em}

\par\vspace{0.6em}\noindent
\begin{minipage}[c]{0.44\linewidth}
  \includegraphics[width=\linewidth]{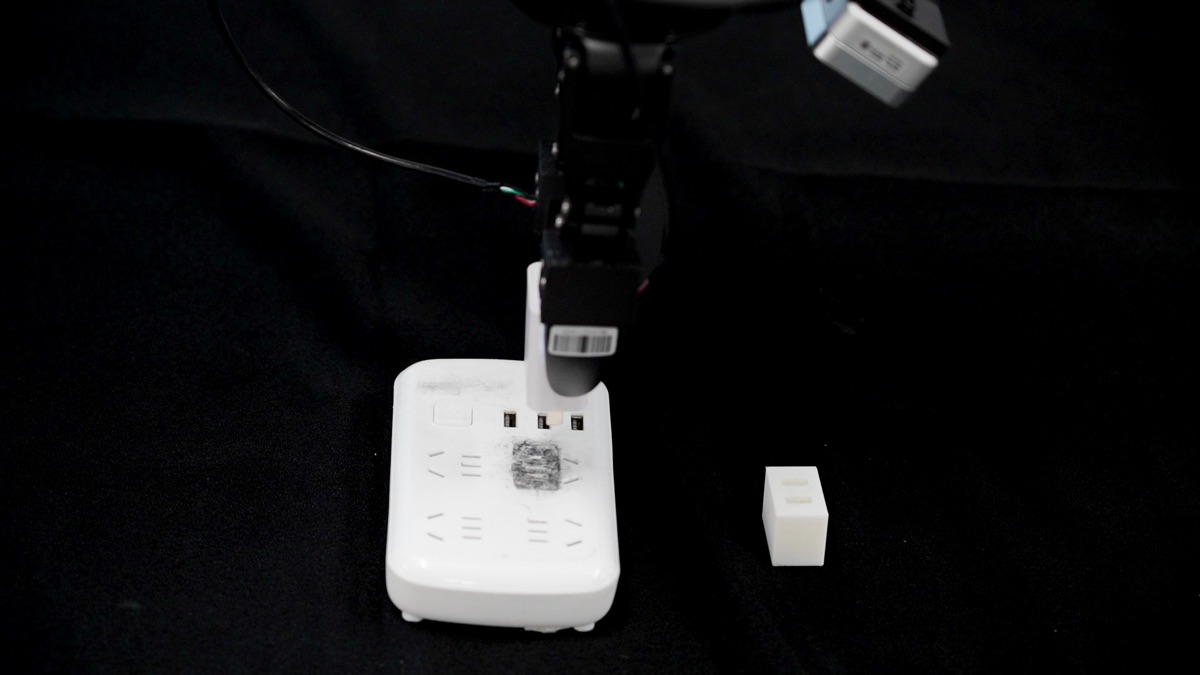}
\end{minipage}\hfill
\begin{minipage}[c]{0.52\linewidth}
  \textbf{Socket Plugging.} The robot grasps a plug, aligns it with the socket, and inserts it along the constrained direction. The task requires tactile detection of edge contact and insertion resistance to avoid jamming or wrong-orientation insertion.
\end{minipage}\par\vspace{0.6em}

\par\vspace{0.6em}\noindent
\begin{minipage}[c]{0.44\linewidth}
  \includegraphics[width=\linewidth]{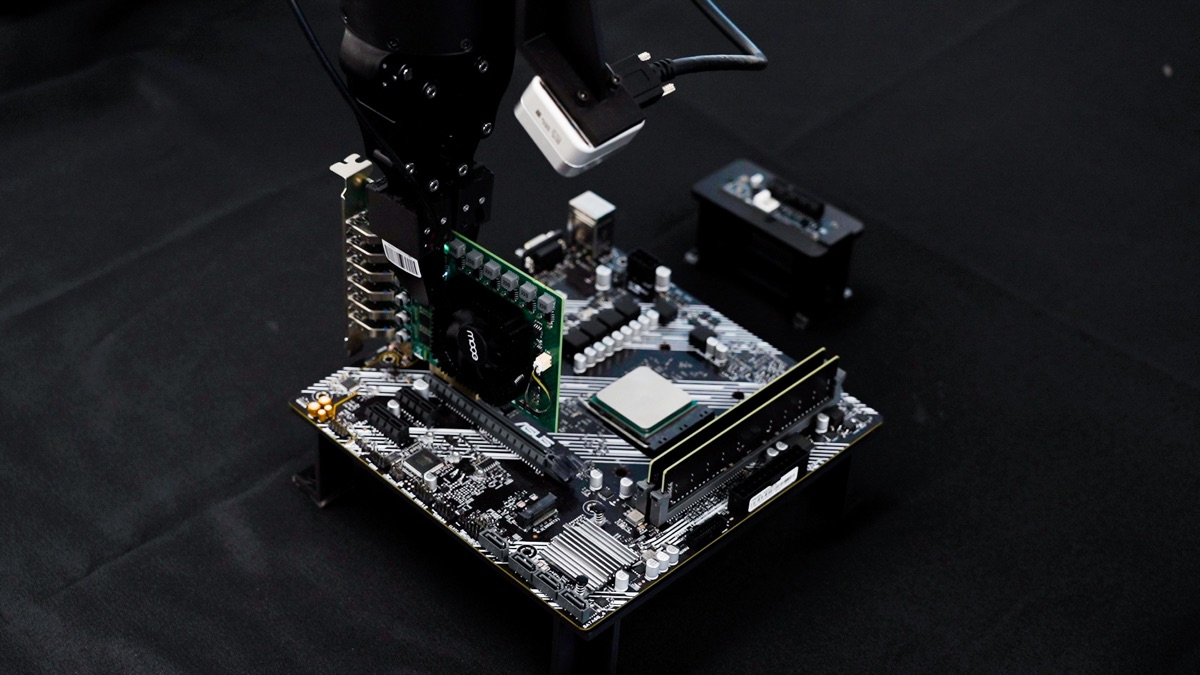}
\end{minipage}\hfill
\begin{minipage}[c]{0.52\linewidth}
  \textbf{Board Insertion.} The robot inserts a circuit board into the target slot or fixture. This precise assembly task stresses small-clearance alignment, low-force insertion, and contact feedback to prevent bending or damage.
\end{minipage}\par\vspace{0.6em}

\par\vspace{0.6em}\noindent
\begin{minipage}[c]{0.44\linewidth}
  \includegraphics[width=\linewidth]{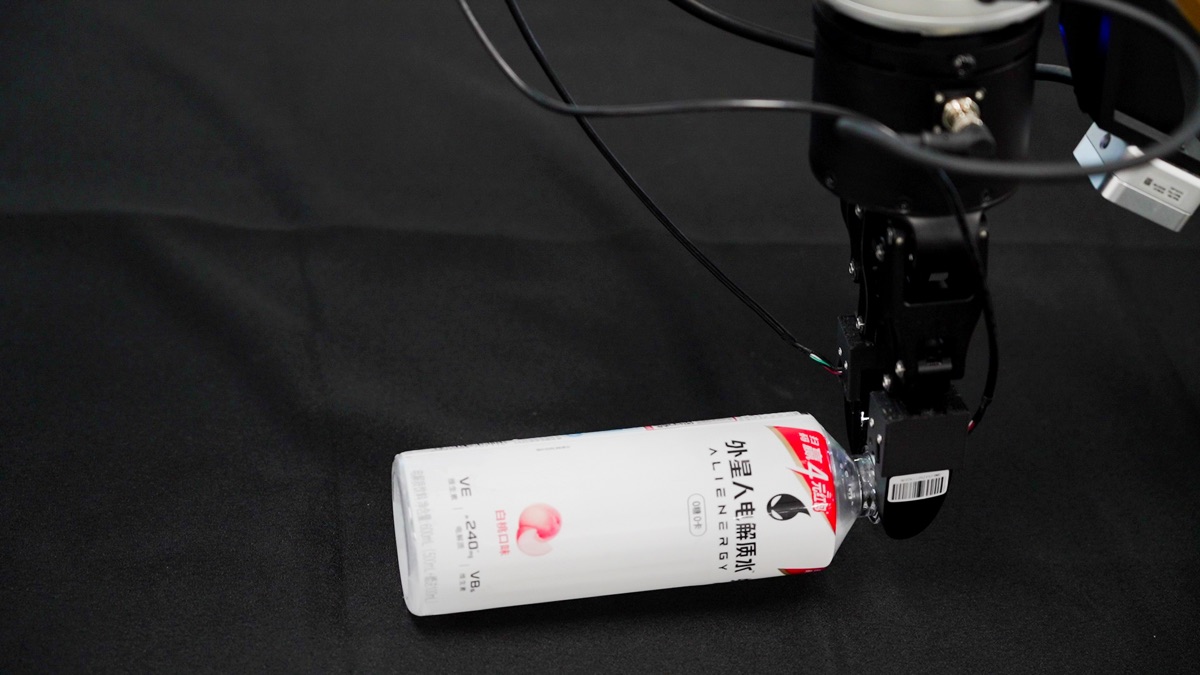}
\end{minipage}\hfill
\begin{minipage}[c]{0.52\linewidth}
  \textbf{Bottle Standing.} The robot recovers a tipped bottle to an upright, stable pose. The task combines curved-surface grasping, center-of-mass management, and controlled release so that the bottle remains standing after the action.
\end{minipage}\par\vspace{0.6em}

\par\vspace{0.6em}\noindent
\begin{minipage}[c]{0.44\linewidth}
  \includegraphics[width=\linewidth]{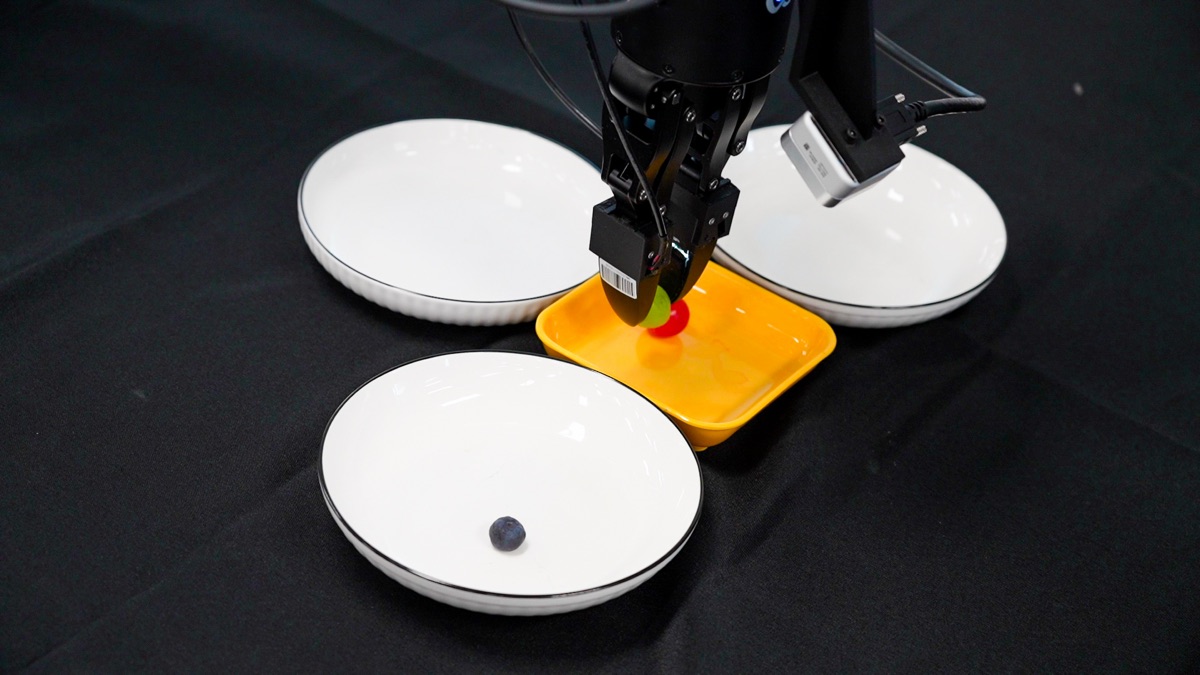}
\end{minipage}\hfill
\begin{minipage}[c]{0.52\linewidth}
  \textbf{Fruit Collection.} The robot picks and places fruit-like objects into the target dish. The task evaluates gentle grasping, transport, and release under a narrow force window that avoids both dropping and visible object damage.
\end{minipage}\par\vspace{0.6em}

\par\vspace{0.6em}\noindent
\begin{minipage}[c]{0.44\linewidth}
  \includegraphics[width=\linewidth]{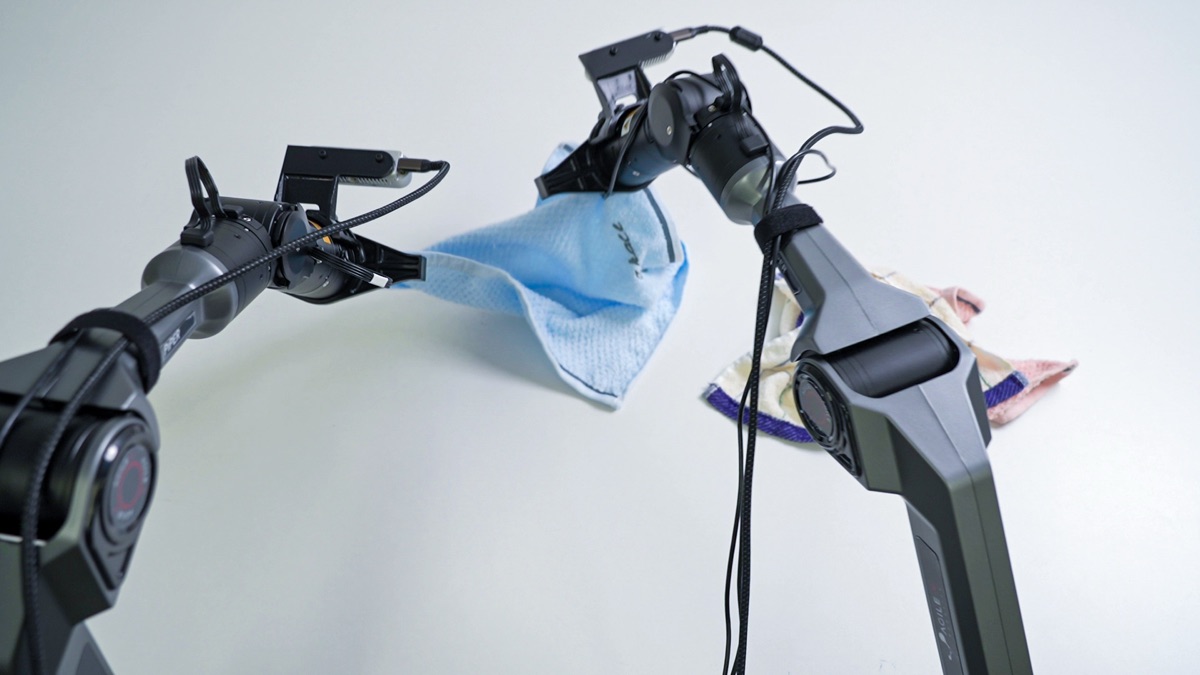}
\end{minipage}\hfill
\begin{minipage}[c]{0.52\linewidth}
  \textbf{Towel Folding.} Two arms manipulate a towel from an unstructured initial state and fold it into the target shape. The task requires deformable-object grasping, edge alignment, flattening, and coordinated bimanual folding.
\end{minipage}\par\vspace{0.6em}

\subsection{NeoReal Task Progressive Score}
\label{sec:neoreal-progressive}

The binary success criterion in Appendix~\ref{sec:neoreal-tasks} scores a trial as $1$ only when the full task goal is reached, which discards all information about how far a policy has progressed before failing. To provide a finer view of partial competence, we additionally define a task progressive score for every NeoReal task. Each task is decomposed into ordered sub-goals that follow its manipulation description, and a fixed number of points is awarded at each key milestone, so that the points accumulate monotonically as the task advances and sum to $100$ at full completion. Tab.~\ref{tab:neoreal-progressive} lists the decomposition and the points assigned to each milestone, one row per task. Unless otherwise noted, a milestone is credited only if all earlier milestones in the same row have been achieved, and the reported progressive score of a trial is the total points of the last milestone reached.

\begin{table}[t]
\centering
\footnotesize
\setlength{\tabcolsep}{4pt}
\renewcommand{\arraystretch}{1.25}
\begin{tabular}{@{}p{0.13\linewidth} p{0.20\linewidth} p{0.20\linewidth} p{0.20\linewidth} p{0.20\linewidth}@{}}
\toprule
Task & Milestone 1 & Milestone 2 & Milestone 3 & Milestone 4 \\
\midrule
Cardboard Box Folding & Grasp and lift the box blank (20) & Fold one side of the box (30) & Press and fold the other side of the box (30) & Box holds the target shape without tearing or collapsing (20) \\
Bag Packing & Grasp the scattered items and put them in the bag (50) & Grasp the zipper pull (20) & Pull the zipper closed along the track (30) & \\
Cable Winding & Grasp the cable (20) & Route it to the first fixture under tension (30) & Wind around the remaining posts along the target path (30) & Complete routing with no missed post, tangle, or crossing (20) \\
Board Wiping & Grasp the wiping tool (15) & Grasp the blackboard to make it stable (25) & Sweep across the marked region while keeping contact (30) & Cover and clean the full target region (30) \\
Cup Stacking & Grasp the first cup (20) & Align it over the second cup (30) & Insert and seat it with limited contact force, no visible deformation (30) & Nested cups form a stable stack (20) \\
Socket Plugging & Grasp the plug (20) & Align it with the socket in the correct orientation (30) & Insert along the constrained direction without jamming (30) & Plug fully seated in the socket (20) \\
Board Insertion & Grasp the circuit board (20) & Align it with the slot at small clearance (30) & Insert with low force and contact feedback (30) & Board fully seated without bending or damage (20) \\
Bottle Standing & Grasp the tipped bottle on its curved surface (25) & Lift and reorient it toward upright (30) & Place it with the center of mass over the base (25) & Release in a controlled manner; bottle stays standing (20) \\
Fruit Collection & Grasp the fruit within the safe force window (30) & Lift without dropping or damaging it (25) & Transport it over the target dish (20) & Release it into the dish (25) \\
Towel Folding & Bimanual grasp of the towel from the initial state (20) & Flatten and spread the towel (25) & Align the target edges (25) & Fold the towel into the target shape (30) \\
\bottomrule
\end{tabular}
\caption{Task progressive score for the ten NeoReal tasks. Each task is decomposed into ordered milestones that follow its description in Appendix~\ref{sec:neoreal-tasks}. The points in parentheses are awarded when the milestone is reached and accumulate along the row, summing to $100$ at full task completion. A milestone is credited only after all earlier milestones in the same row have been achieved, so the progressive score of a trial equals the total points of the last milestone it reaches.}
\label{tab:neoreal-progressive}
\end{table}

\paragraph{Per-execution scoring.}
Following Tab.~\ref{tab:neoreal-progressive}, every rollout is scored by the furthest milestone it reaches. We walk through the ordered milestones of the task and accumulate their points until the execution stops making progress or triggers one of the failure conditions in Section~\ref{sec:neoreal-eval}, and the progressive score of that rollout is the total accumulated points. A rollout that completes the task passes all milestones and therefore scores the full $100$ points, whereas a rollout that fails partway is credited only for the milestones it has already achieved. The per-task progressive score reported in Fig.~\ref{fig:neoreal-results} is the mean of these per-rollout scores over the $20$ evaluation trials. A successful trial always attains all milestones and contributes the maximum $100$ points, while partially completed trials still contribute positive points, so the progressive score of a task is always at least as large as its binary success rate and exposes the partial competence that the success rate alone hides.

\subsection{NeoReal Training Settings}

NeoReal policies follow a pretrain then post train recipe. In the pretraining stage, each policy is pretrained on over $400{,}000$ episodes from NeoData to acquire broad contact-rich manipulation priors. In the post-training stage, the pretrained policy is specialized to each NeoReal task using $300$ demonstrations that are collected specifically for that task.

All policies and baselines are trained on $8\times$ NVIDIA A100 GPUs. Both the third-person external view and the wrist view are provided as visual inputs, and actions are parameterized as end-effector poses for all tasks. Single-arm tasks are executed on a Flexiv Rizon 4s arm, while dual-arm tasks are executed on Piper or ARX X5 arms.

\subsection{NeoReal Evaluation Settings}
\label{sec:neoreal-eval}

Each task is evaluated over $20$ trials. At the start of every trial we reset the robot arm to an initial pose sampled within a bounded range, so that the evaluation carries a controlled degree of randomization and tests generalization over the starting configuration. The positions of the manipulated assets are likewise randomized within their own ranges, adding spatial generalization to the object layout.

During a trial, we count the run as an execution failure if the arm reaches a joint limit, moves out of the workspace boundary, enters a self-locking configuration, or becomes stuck against or collides with the environment. If the arm repeats the same action more than $5$ times without progress, we count it as a task progression failure. If the task is not completed within $5$ minutes, we count it as a timeout failure. A trial is scored as a success only when the arm completes all milestones of the task defined in the task progressive score of Appendix~\ref{sec:neoreal-progressive} without triggering any of the above failures. We report the mean success rate across the $20$ trials.

\section{NeoSim Task Suite}
\label{sec:neosim-tasks}

This appendix describes the NeoSim benchmark, covering its simulation platform, the $12$ contact-rich tasks ($4$ single-arm and $8$ dual-arm), and the training and evaluation settings. For each task we show a rendered example, with the simulated RGB scene on the left and the corresponding particle gel images on the right following the format of Fig.~\ref{fig:benchmark-neosim-demonstration}, and describe the manipulation skill it exercises and the role that tactile feedback plays.

\subsection{Simulation Platform}
\label{sec:neosim-platform}

NeoSim is built on UniVTAC~\citep{chen2026univtac}, a visuo-tactile simulation platform based on Isaac Sim and Isaac Lab~\citep{mittal2023orbit} whose tactile layer is provided by TacEx~\citep{nguyen2024tacex}. The sensor gel is simulated as a finite element soft body with incremental potential contact and outputs tactile RGB images, marker motion, and gel depth maps. On top of this platform, we additionally simulate a markerless particle gel surface, where a dense speckle coating is advected by the finite element surface displacement, whose tangential component is taken from marker motion and whose normal component is taken from the gel depth map. We also derive per-vertex contact forces from the contact solver and interpolate them into the dense force field that serves as the tactile ground truth in NeoSim.

Each task is instantiated from a modular configuration file that specifies the task assets, the initial object layout, the randomization ranges, and the recorded observation streams. Initial object poses are randomized per episode within task-specific ranges. Success is judged by task-specific geometric criteria. For insertion tasks, for example, a trial counts as successful only when the axial alignment, the radial offset, and the insertion depth all reach their thresholds.

\subsection{Single-arm Tasks}

The four single-arm tasks are executed with a single arm and stress force regulation, insertion precision, and damage-free grasping.

\par\vspace{0.6em}\noindent
\begin{minipage}[c]{0.44\linewidth}
  \includegraphics[width=\linewidth]{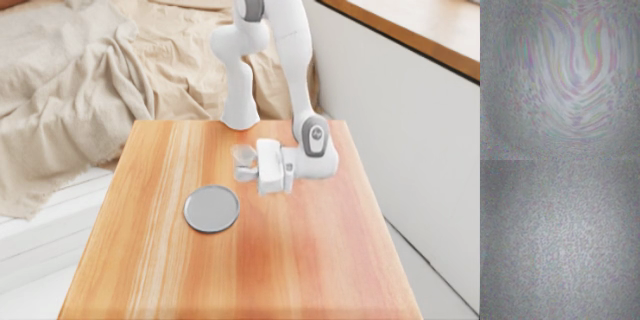}
\end{minipage}\hfill
\begin{minipage}[c]{0.52\linewidth}
  \textbf{Pour Ball.} The arm grasps a cup filled with small balls and pours them onto a target plate. The cup is compliant and the balls shift during motion, so the policy must maintain a stable grasp force and modulate the pouring angle, and the particle gel rendering reveals how the contents load the fingers, which vision alone cannot observe.
\end{minipage}\par\vspace{0.6em}

\par\vspace{0.6em}\noindent
\begin{minipage}[c]{0.44\linewidth}
  \includegraphics[width=\linewidth]{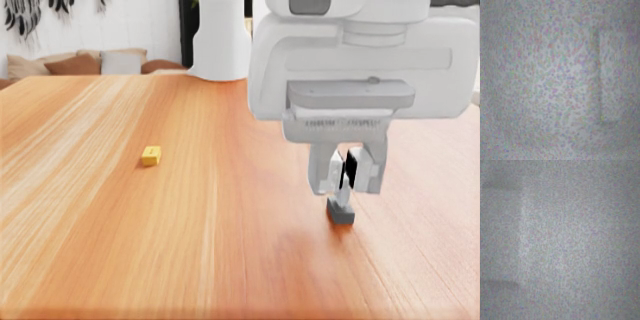}
\end{minipage}\hfill
\begin{minipage}[c]{0.52\linewidth}
  \textbf{Unplug and Plug Charger.} The arm disconnects a charger connector from a socket and inserts it again. Both the release and the insertion depend on sensing the seating force and the small resistance at the moment of connection, making this a canonical force-guided mating task.
\end{minipage}\par\vspace{0.6em}

\par\vspace{0.6em}\noindent
\begin{minipage}[c]{0.44\linewidth}
  \includegraphics[width=\linewidth]{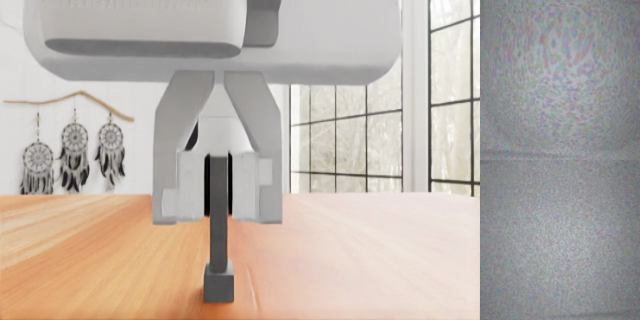}
\end{minipage}\hfill
\begin{minipage}[c]{0.52\linewidth}
  \textbf{Plug USB.} A precision peg-in-hole task in which the arm aligns and inserts a USB connector into its port. The tight-clearance means that success hinges on tactile detection of edge contact and small misalignment before the connector jams.
\end{minipage}\par\vspace{0.6em}

\par\vspace{0.6em}\noindent
\begin{minipage}[c]{0.44\linewidth}
  \includegraphics[width=\linewidth]{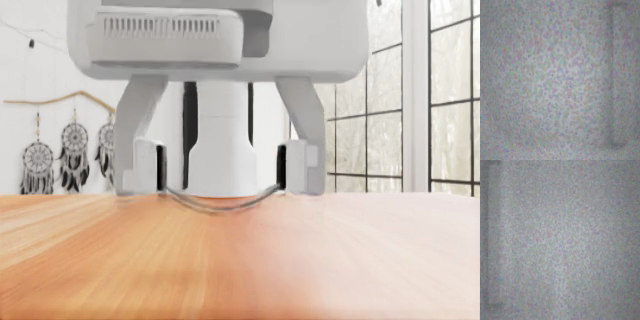}
\end{minipage}\hfill
\begin{minipage}[c]{0.52\linewidth}
  \textbf{Grasp Chip.} The arm grasps a thin, fragile chip and lifts it without crushing or dropping it. This task defines a narrow force window, since too little force lets the chip slip while too much force damages it, so it directly probes fine-grained force limiting.
\end{minipage}\par\vspace{0.6em}

\subsection{Dual-Arm Tasks}

The eight dual-arm tasks require two arms and additionally test bimanual coordination and stable contact during handover, insertion, and stacking.

\par\vspace{0.6em}\noindent
\begin{minipage}[c]{0.44\linewidth}
  \includegraphics[width=\linewidth]{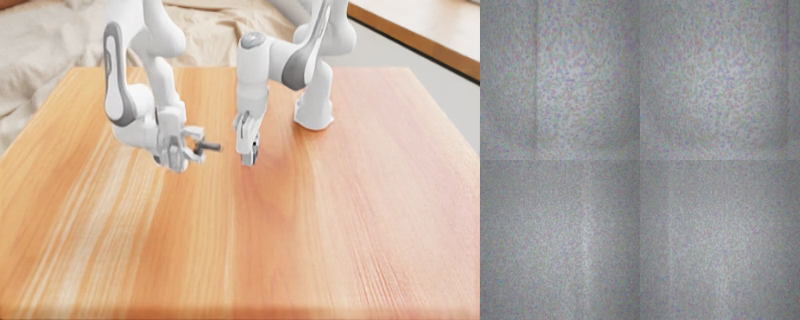}
\end{minipage}\hfill
\begin{minipage}[c]{0.52\linewidth}
  \textbf{Insert Screw.} One arm stabilizes a sleeve while the other inserts and threads a screw into it. Coordinated force sensing on both arms is needed to keep the parts aligned and to detect proper seating during threading.
\end{minipage}\par\vspace{0.6em}

\par\vspace{0.6em}\noindent
\begin{minipage}[c]{0.44\linewidth}
  \includegraphics[width=\linewidth]{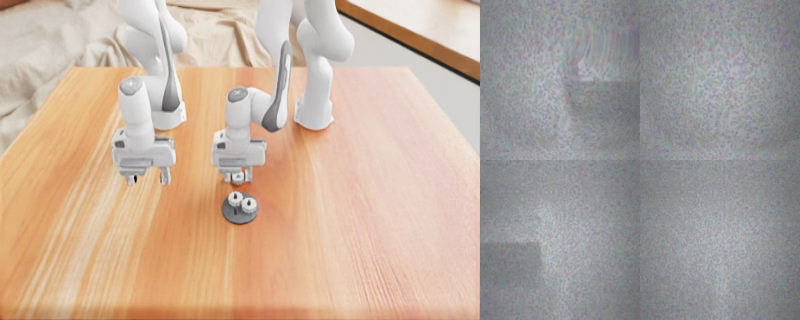}
\end{minipage}\hfill
\begin{minipage}[c]{0.52\linewidth}
  \textbf{Place Gears.} The arms pick up gears and thread them onto holder pegs. Tactile feedback confirms that each gear is seated flush against the peg rather than resting on an edge, a state that is visually ambiguous.
\end{minipage}\par\vspace{0.6em}

\par\vspace{0.6em}\noindent
\begin{minipage}[c]{0.44\linewidth}
  \includegraphics[width=\linewidth]{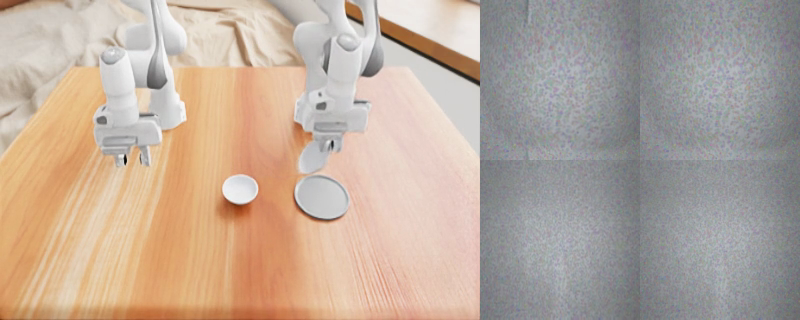}
\end{minipage}\hfill
\begin{minipage}[c]{0.52\linewidth}
  \textbf{Unstack Bowl.} One arm grips the rim of the top bowl in a nested stack and lifts it clear while the other stabilizes the remaining stack. Detecting the thin rim through touch is decisive, since the contact surface is small and easily missed by vision.
\end{minipage}\par\vspace{0.6em}

\par\vspace{0.6em}\noindent
\begin{minipage}[c]{0.44\linewidth}
  \includegraphics[width=\linewidth]{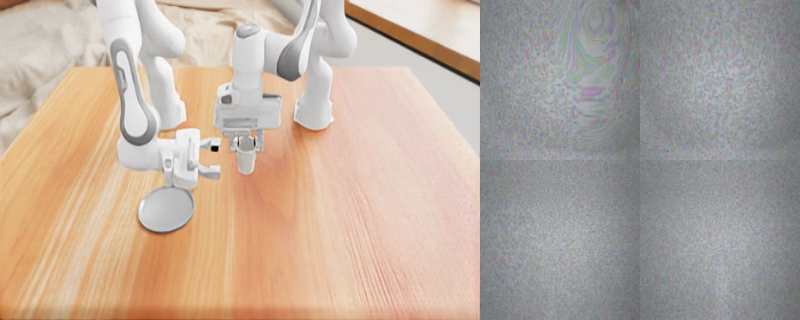}
\end{minipage}\hfill
\begin{minipage}[c]{0.52\linewidth}
  \textbf{Cup Handover.} One arm grasps a cup and hands it to the second arm, which then places it. The handover moment requires both arms to sense the shared contact so that the cup is neither dropped nor crushed during the transfer.
\end{minipage}\par\vspace{0.6em}

\par\vspace{0.6em}\noindent
\begin{minipage}[c]{0.44\linewidth}
  \includegraphics[width=\linewidth]{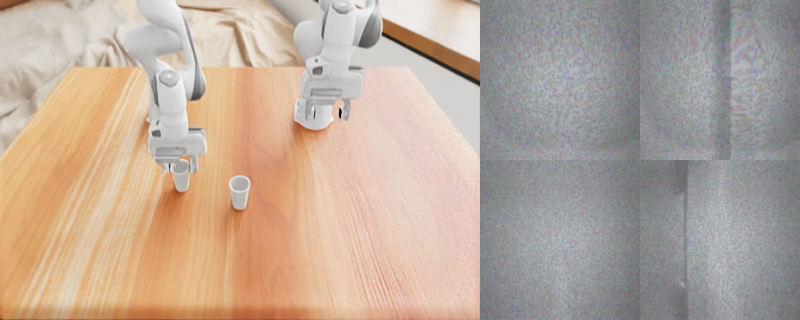}
\end{minipage}\hfill
\begin{minipage}[c]{0.52\linewidth}
  \textbf{Stack Cups.} Two separated cups are brought together and nested into a centered stack. The policy must align the cups and sense the seating contact as one cup enters the other.
\end{minipage}\par\vspace{0.6em}

\par\vspace{0.6em}\noindent
\begin{minipage}[c]{0.44\linewidth}
  \includegraphics[width=\linewidth]{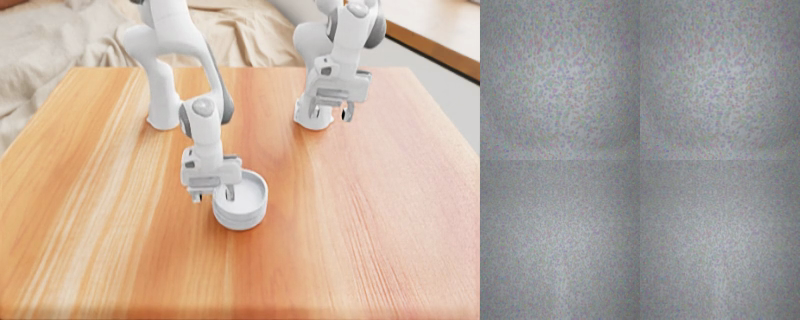}
\end{minipage}\hfill
\begin{minipage}[c]{0.52\linewidth}
  \textbf{Stack Plates.} Two separated plates are combined into a centered stack. The thin, flat geometry makes tactile confirmation of centered, flush contact important for a stable stack.
\end{minipage}\par\vspace{0.6em}

\par\vspace{0.6em}\noindent
\begin{minipage}[c]{0.44\linewidth}
  \includegraphics[width=\linewidth]{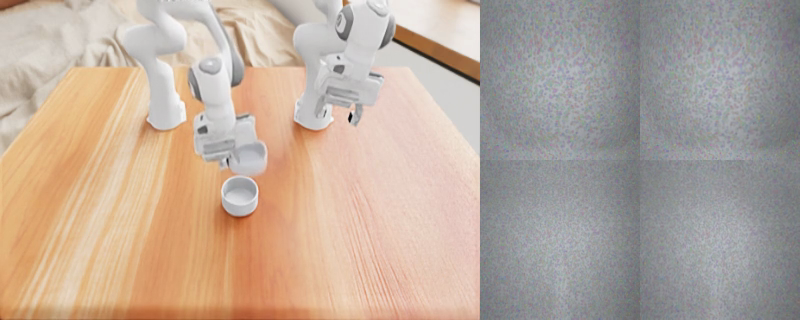}
\end{minipage}\hfill
\begin{minipage}[c]{0.52\linewidth}
  \textbf{Stack Bowls.} Two separated bowls are brought together for centered nesting. As with plates, tactile alignment ensures the bowls seat concentrically rather than off-center.
\end{minipage}\par\vspace{0.6em}

\par\vspace{0.6em}\noindent
\begin{minipage}[c]{0.44\linewidth}
  \includegraphics[width=\linewidth]{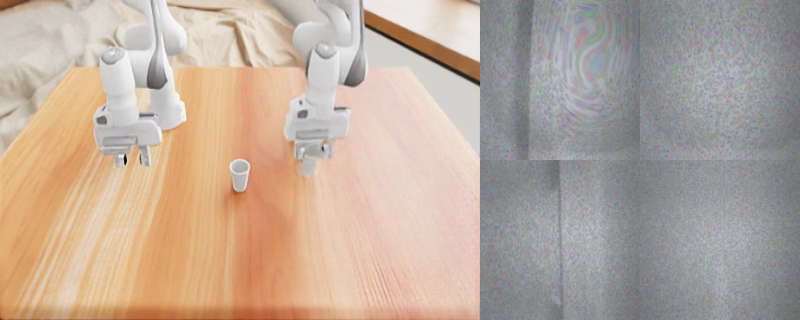}
\end{minipage}\hfill
\begin{minipage}[c]{0.52\linewidth}
  \textbf{Unstack Cup.} One arm manipulates the top cup of a nested cup pair while the other stabilizes the base. The tight fit of nested cups increases the separation force the policy must sense and overcome, mirroring Unstack Bowl at a smaller scale.
\end{minipage}\par\vspace{0.6em}

\subsection{NeoSim Training Settings}
\label{sec:neosim-training}

Demonstrations are collected automatically by scripted expert policies built on the cuRobo motion planner~\citep{sundaralingam2023curobo}. An episode is kept only when the planner succeeds and the final state passes the geometric success check, and collection continues until $100$ successful episodes are obtained per task. All evaluated policies are trained on the same $100$ demonstrations for each task. Every raw episode stores synchronized external and wrist RGB views, the two finger tactile streams, and both joint and end-effector states, so the demonstrations can be converted into the native input format of each policy, such as joint space actions for some baselines and end-effector poses combined with joint states for others. Each task-and-policy combination is trained on $4\times$ NVIDIA A100 GPUs.

\subsection{NeoSim Evaluation Settings}
\label{sec:neosim-eval}

Each task is evaluated over $100$ rollouts generated from $100$ fixed random seeds that are disjoint from the seeds used for demonstration collection, so all policies face the same set of initial configurations. A rollout terminates when the task succeeds, when an early stop condition is triggered, or when the step limit is reached, and this limit is set per-task according to its difficulty. The reported metric is the mean success rate over the $100$ rollouts. Policy inference runs on NVIDIA RTX 4090 GPUs and communicates with the simulator through a remote inference interface, which decouples policy serving from simulation and allows evaluations to be deployed across multiple machines under an identical environment configuration.

\end{document}